\documentclass[runningheads]{llncs}

\usepackage{eccv}

\usepackage{eccvabbrv}

\usepackage{xcolor}
\usepackage{booktabs}       
\usepackage{nicefrac}       
\usepackage{microtype}      
\usepackage{graphicx}   
\usepackage{doi}
\usepackage{wrapfig}
\usepackage{float} 
\usepackage{import}
\usepackage{caption}
\usepackage{subcaption} 
\usepackage{adjustbox}
\usepackage{comment}
\usepackage{acronym}
\usepackage{xspace}
\usepackage{array}
\usepackage[mathscr]{eucal}
\usepackage{multirow}
\usepackage{amsmath,amssymb,amsfonts}
\usepackage{multicol}
\usepackage{algorithm}
\usepackage{algpseudocode}
\usepackage{changepage}
\usepackage{enumitem}
\usepackage{lscape}

\acrodef{ML} 		[\textsc{ML\xspace}]				{Machine Learning}
\acrodef{DL} 		[\textsc{DL\xspace}]				{Deep Learning}
\acrodef{ANN} 		[\textsc{ANN\xspace}]				{Artificial Neural Network}
\acrodef{DNN} 		[\textsc{DNN\xspace}]				{Deep Neural Network}
\acrodef{DNNs} 		[\textsc{DNNs\xspace}]				{Deep Neural Networks}
\acrodef{CNNs} 		[\textsc{CNNs\xspace}]				{Convolutional Neural Networks}
\acrodef{CNN} 		[\textsc{CNN\xspace}]				{Convolutional Neural Network}
\acrodef{MTL} 		[\textsc{MTL\xspace}]				{Multi-Task Learning}
\acrodef{STL} 		[\textsc{STL\xspace}]				{Single-Task Learning}
\acrodef{RL} 		[\textsc{RL\xspace}]				{Reinforcement Learning}
\acrodef{LL} 		[\textsc{LL\xspace}]				{Lifelong Learning}
\acrodef{NLP} 		[\textsc{NLP\xspace}]				{Natural Language Processing}
\acrodef{LSTM} 		[\textsc{LSTM\xspace}]				{Long Short-Term Memory}
\acrodef{MAML} 		[\textsc{MAML\xspace}]				{Model Agnostic Meta Learning}
\acrodef{GAN}   	[\textsc{GANs\xspace}]	            {Generative Adversarial Nets}
\acrodef{VAE}   	[\textsc{VAEs\xspace}]	            {Variational Auto-Encoders}
\acrodef{NAS}   	[\textsc{NAS\xspace}]	            {Neural Architecture Search}
\acrodef{LOMT}   	[\textsc{LOMT\xspace}]	            {Layer-optimized Multi-Task}

\usepackage[accsupp]{axessibility}  

\usepackage{hyperref}

\usepackage{orcidlink}

\begin{document}

\title{Giving each task what it needs - leveraging structured sparsity for tailored multi-task learning} 

\titlerunning{Giving each task what it needs}

\author{Richa Upadhyay\inst{1}\orcidlink{0000-0001-9604-7193}\and
Ronald Phlypo\inst{2}\orcidlink{0000-0003-4310-7994}\and
Rajkumar Saini\inst{1}\orcidlink{0000-0001-8532-0895}\and
Marcus Liwicki\inst{1}\orcidlink{0000-0003-4029-6574}
}

\authorrunning{Upadhyay et al.}

\institute{Lule\aa~ tekniska universitet, Lule\aa, Sweden\\  
\email{firstname.lastname@ltu.se}\\ 
\and
University Grenoble Alpes, Grenoble, France\\
\email{ronald.phlypo@grenoble-inp.fr}}

\maketitle

\begin{abstract}
 In the Multi-task Learning (MTL) framework, every task demands distinct feature representations, ranging from low-level to high-level attributes.
It is vital to address the specific (feature/parameter) needs of each task, especially in computationally constrained environments.
This work, therefore, introduces Layer-Optimized Multi-Task (LOMT) models that utilize structured sparsity to refine feature selection for individual tasks and enhance the performance of all tasks in a multi-task scenario. 
Structured or group sparsity systematically eliminates parameters from trivial channels and, sometimes, eventually, entire layers within a convolution neural network during training.
Consequently, the remaining layers provide the most optimal features for a given task.
In this two-step approach, we subsequently leverage this sparsity-induced optimal layer information to build the LOMT models by connecting task-specific decoders to these strategically identified layers, deviating from conventional approaches that uniformly connect decoders at the end of the network. 
This tailored architecture optimizes the network, focusing on essential features while reducing redundancy.
We validate the efficacy of the proposed approach on two datasets, \ie NYU-v2 and CelebAMask-HD datasets, for multiple heterogeneous tasks. 
A detailed performance analysis of the LOMT models, in contrast to the conventional MTL models, reveals that the LOMT models outperform for most task combinations. 
The excellent qualitative and quantitative outcomes highlight the effectiveness of employing structured sparsity for optimal layer (or feature) selection. 
\end{abstract}

\section{Introduction} \label{sec:intro}
\ac{MTL} represents a learning paradigm where multiple related tasks can be learned jointly, facilitating the sharing of features or parameters among the tasks to enhance their mutual learning \cite{caruana1997multitask}.
This cooperative learning can be established in primarily two ways: hard parameter sharing, which is a result of the inherent design of multi-task architectures, and soft parameter sharing, which arises from constraints applied to the parameters \cite{crawshaw2020multi}.
This work is centered around hard parameter sharing, wherein the multi-task architecture explicitly shares the common backbone (encoder/feature extraction layers) parameters between tasks. 
Within a multi-task framework, the distinct feature requirements of various tasks highlight the significance of task-specific feature selection.
Overlooking this critical aspect may result in suboptimal model performance, emphasizing the need for precise feature curation, tailored to the specific requirements of each task. 
Therefore, this work exploits structured (group) sparsity techniques for task-specific feature selection.

\begin{figure}[t]
    \centering
    \includegraphics[width = 0.8\linewidth]{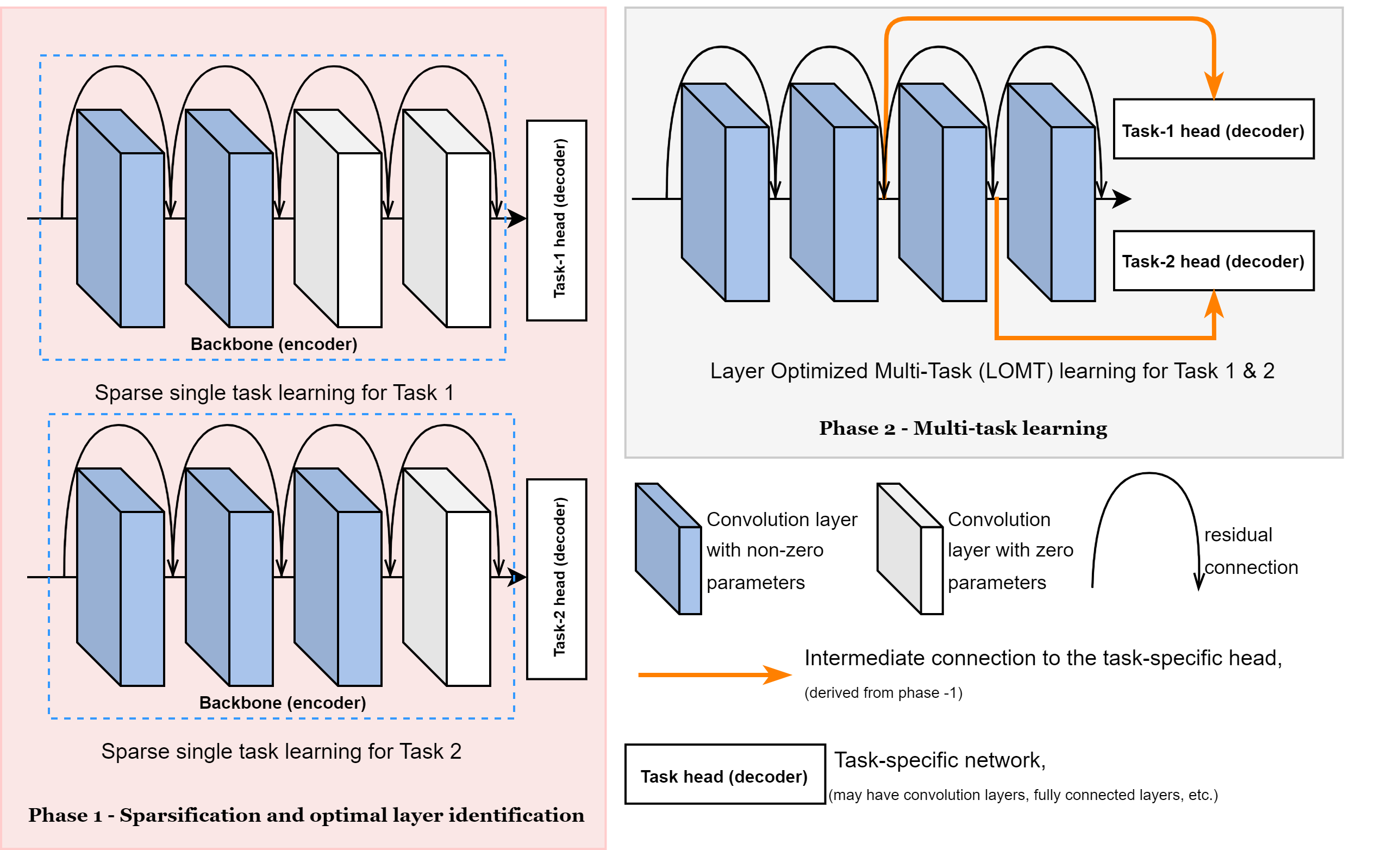}
    \caption{A simple schematic representation of the proposed work.}
    \label{fig:block_diag}
\end{figure}

While model sparsification is typically employed for model compression by eliminating a portion of extraneous model parameters \cite{review_sparsity}, in this study, we leverage this property for feature selection, more specifically, layer selection, for training efficient multi-task models.
In the context of \ac{DL}, sparsity frequently indicates a model's capacity to effectively generalize from the data by identifying and utilizing the most prominent features.
In this work, we use group sparsity in the backbone(encoder) network of each task individually \ie during \ac{STL}.
This helps to identify the features or layers that are crucial for a task while eliminating (setting to zero) redundant features that eventually do not help improve the task's performance.
This knowledge of essential features in terms of non-sparse \ac{CNN} layers during task-specific learning is then utilized during \ac{MTL}, as shown in Fig.~\ref{fig:block_diag}. 
The optimal layer information for each task is shared from the sparse single-task models in phase 1 to create \ac{LOMT} models in phase 2.

Overall, the core idea behind \ac{LOMT} models is to leverage sparsity-induced feature selection for individual tasks in a way that preserves only the most relevant and essential information for those tasks.
By connecting task-specific decoders to their corresponding, optimally identified intermediate layers rather than the final layer of a shared encoder, LOMT models aim to tailor the network to the distinctive requirements of each task, encouraging more focused and efficient learning. 
The main contributions of this work are two-fold:
\noindent
\begin{enumerate}
    \item This work presents a straightforward two-step approach for implementing optimal \ac{MTL} by employing structured sparsity to determine and utilize the most significant layers for diverse tasks, thereby enhancing model performance.
    \item We present a comprehensive comparative performance analysis of the proposed work in contrast to the dense \ac{STL} and \ac{MTL} for two datasets and on many heterogeneous tasks, which involve pixel-level dense prediction tasks and image-level classification tasks. We also provide insights into the quantitative results and explainability of the \ac{LOMT} models using activation maps. 
\end{enumerate}

\section{Related work} \label{sec:related_work}
Within the field of Multi-Task Learning (MTL), research primarily focuses on three main areas: (i) the methodology for training tasks together (how to train the tasks together?), with a specific emphasis on developing complex multi-task architectures, as in \cite{liu2019end}, \cite{misra2016cross}, \cite{gao2019nddr}, \cite{vandenhende2020mti}, \cite{lu2017fully}, \cite{kanakis2020reparameterizing} and many more, (ii) choosing the group of tasks to be trained simultaneously (which tasks to train together?), with a particular focus on analyzing task correlation, and transference, like \cite{standley2020tasks}, \cite{Zamir_2018_CVPR}, \cite{guo2018dynamic}, \cite{kumar2012learning}, among others, and (iii) identifying the features or information that should be shared among tasks (what to share between the tasks?) as discussed in \cite{sun2020adashare}, \cite{ruder2019latent}, \cite{vandenhende2020branched}, \cite{guo2020learning}, \cite{bruggemann2020automated}, \cite{choi2023dynamic}, \cite{paper3}, and many more.
Since this work focuses on feature selection in a \ac{MTL} setting for various tasks, we will limit our discussion to the third area, which emphasizes effective feature sharing between the tasks.

\cite{sun2020adashare} introduces AdaShare, an adaptive sharing technique to optimize recognition accuracy and resource efficiency by creating a task-specific policy that determines the most effective layers to activate for a particular task within a multi-task setting. 
Learning AdaShare's effective sharing strategy requires understanding both tasks and their interactions, which may increase the training time and computational resources required.
\cite{guo2020learning} employs the concepts of \ac{NAS} to automatically create hard parameter-sharing multi-task networks by utilizing the update gradients derived from the combined task losses.
It eliminates the need for pre-defined task-relatedness scores and alters the network branches to produce network configurations that are both effective and efficient. 
Similarly, \cite{bruggemann2020automated} and \cite{choi2023dynamic} are also inspired by \ac{NAS}.
The former uses a resource-aware loss that adaptively controls the network size while creating tree-like structures in the encoder of a multi-task network. In contrast, the latter uses Directed Acyclic Graphs (DAG) to search for richer network typologies for a multi-task problem.
The use of \ac{NAS} for developing efficient multi-task architectures has been extensively studied in the literature; however, this approach is often associated with high computational costs.
\cite{vandenhende2020branched}, on the other hand, overcomes this issue by leveraging task affinities.
They use the representation similarity analysis to find the task affinities at a few set locations in the neural network, which is further used to construct branched multi-task networks autonomously.
A meta-network is introduced in \cite{ruder2019latent} that connects the single-task learning networks in the latent feature space and learns what features are to be shared between the tasks. 
It not only determines the specific layers to be shared among tasks and the extent to which they are shared but also integrates a mixed model of skip connections at its outermost layer.
This work is like an expansion of the various \ac{MTL} algorithms.
The paper \cite{paper3} is a highly relevant and comparable study to our current work. 
It employs group sparsity for feature selection in multi-task settings. In contrast, we utilize it in single-task settings and subsequently leverage the acquired information for \ac{MTL}.
In this study, we compared performance with \cite{paper3}.
In addition, we have compared our work to a few others that have utilized the same dataset and performed the same tasks.

\textit{Positioning our work}- Unlike the previous research that mainly uses advanced \ac{NAS} approaches, reinforcement learning strategies, or different feature selection methods for multi-task learning, this study presents a more straightforward and intuitive approach.
This work primarily aims to determine the most relevant subset of \ac{CNN} layers most suitable for each task when trained in isolation \ie single-task learning. 
We identify the most crucial layers for each task by applying the channel-wise group sparsity across all the convolution layers of the \ac{CNN}.
This focused application of penalty-based sparsity techniques systematically prunes (set to zero) the less significant features, thereby highlighting the important layers for achieving better task-specific performance. 
This vital optimal layer information is subsequently utilized within a multi-task framework.

\section{Methodology} \label{sec:method}
In this section, we develop the methodology of the proposed work in two phases. 
The first phase entails single-task learning combined with group sparsity to minimize unnecessary parameters for a specific task. 
The next phase utilizes the knowledge of non-redundant layers or features obtained from the previous phase for multi-task learning.

\textit{Phase -1}: Model sparsification has always been used widely in the literature to eliminate redundant model parameters by setting a considerable number of parameters to zero. 
The two main variants of how sparsity can be induced into the model parameters are unstructured and structured sparsity \cite{review_sparsity}. 
Unstructured sparsity involves zeroing parameters without following any predefined pattern. 
However, structured sparsity, also known as group sparsity, zeroes out predefined groups of parameters. 
Since the parameter vectors of a convolutional layer are naturally organized into groups like filters, channels, and so on, it provides an excellent opportunity for implementing structured sparsity.
This work centers on utilizing structured sparsity techniques to isolate parameters that are essential for a specific task.
In line with the previous related works by \cite{wen2016learning}, \cite{deleu2021structured}, \cite{paper3}, and many more, we selectively refine the model parameters by applying channel-wise group sparsity to the convolution layers of the encoder, also known as the backbone of the network during \ac{STL}.
Sparsification is limited to the encoder to ensure that the resulting sparse layer insights can be efficiently utilized during \ac{MTL} of the given task.

\begin{figure}
\centering
\begin{subfigure}[b]{\textwidth}
   \includegraphics[width=1\linewidth]{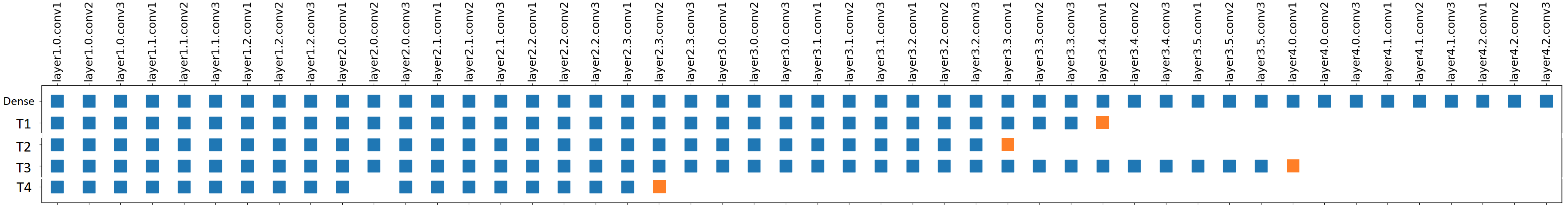}
   \caption{NYU-v2 dataset}
\end{subfigure}
\begin{subfigure}[b]{\textwidth}
   \includegraphics[width=1\linewidth]{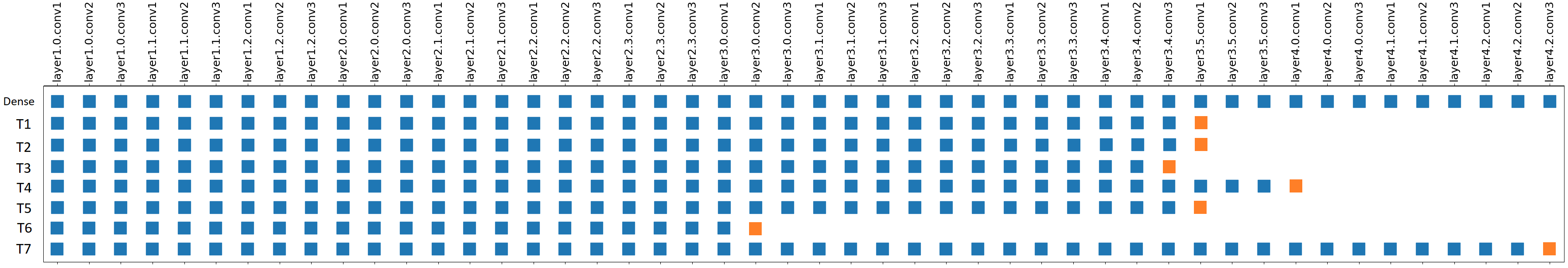}
   \caption{CelebAMask-HQ dataset}
\end{subfigure}
\caption{Sparsity pattern of the layers of ResNet-50 backbone for (a) NYU-v2 and (b) CelebAMask-HQ dataset. The first row represents the dense network, with all layers having non-zero parameters indicated in blue. The subsequent rows illustrate the sparsity pattern obtained after training a single-task model using group sparsity. The orange square stands for the last layer, after which all the parameters are zero, and this layer is selected for phase 2 to create \ac{LOMT} models. The names of tasks $T_n$ for both the figures can be found in Tables.~\ref{tab:NYU_results_res} and \ref{tab:celebA_results}.}
\label{fig:pattern}
\end{figure}

Penalty or regularization-based $l_1l_2$ group sparsity\cite{yuan2006model} is employed to achieve structural sparsity in the network \cite{deleu2021structured}.
The optimization objective now is to minimize a combined function of loss $\mathcal{L}$ and the $l_1l_2$ penalty term $\mathcal{R}$ given by the Eq.~\ref{eq:sp_obj},
\begin{equation}\label{eq:sp_obj}
    \min_{\theta = \{\theta_b,\theta_t\}} \mathcal{L}(\theta) + \mathcal{R}(\theta_b), ~~ where, ~~ \mathcal{R} = \sum_{g = 1}^{G} \lambda_g ||\theta_b^g||_2
\end{equation}
here $\theta$ represent the model parameters, such that $\theta = \{\theta_b, \theta_t\}$, with $\theta_b$ being the backbone (shared) network parameters and $\theta_t$ the task-specific (head or decoder) parameters.
For any group $g$, $\lambda_g = \lambda \sqrt{N_g}$, where $\lambda$ is the sparsity parameter and $N_g$ stands for the number of elements in group $g$. 
G is the total number of groups, in this case, the total number of channels across all the convolution layers of the backbone network.
The penalty term $\mathcal{R}$ is the $l_1$ norm of the $l_2$ norm of the groups of parameters, which is why it is also known as $l_1l_2$ sparsity.
Since the penalty term makes the objective function non-differentiable, we employ proximal gradient methods for optimizing the combined function in Eq.\ref{eq:sp_obj}, discussed in detail in \cite{combettes2005signal}, \cite{o2016statistical}, \cite{deleu2021structured}.
The proximal gradient updates are based on the gradient updates of the differential part \ie $\mathcal{L}(\theta)$ given by Eq.~\ref{prox1},
\begin{equation}\label{prox1}
    \theta_{n+1} \leftarrow prox_{\alpha \mathcal{R}}(\theta_n - \alpha \nabla_{\theta}\mathcal{L}(\theta_n)),~~~ here~n~stands~for~the~n^{th}~update.
\end{equation}
For the penalty-based $l_1l_2$ group sparsity, the above equation has a closed form given by,
\begin{equation}
    prox_{\alpha \mathcal{R}}(\theta^g) = \begin{cases}\left[1- \dfrac{\alpha \lambda_g}{||\theta^g||_2}\right]\theta^g & ;||\theta^g||_2 \geq \lambda_g\\0 &  ;||\theta^g||_2 < \lambda_g\end{cases}
\end{equation}
So, while training the network with group sparsity, entire groups (channels) of parameters progressively exhibit sparsity if the norm of the group of parameters is less than the sparsity parameter of the group \ie $\lambda_g$.
Instead of imposing a sparsity budget, such as expecting x\% parameter sparsity in the network, we let the network adapt dynamic sparsity patterns while training.  
Also, in this work, we use a dilated ResNet50 \cite{yu2017dilated} as the backbone network so that the residual connections help to propagate the values ahead even if the parameters of an entire layer are zeroed out. 

We get a sparsity pattern for each task with many channels reduced to zero, as in Fig.~\ref{fig:pattern}.
We derive the information about the layer until which all the parameters are mostly non-zero (\eg the layers represented by the orange square in Fig.~\ref{fig:pattern}) and after which the subsequent layers are set to zero because of group sparsity. 
Note that in the non-zero layers, a few channels may exist that are zeroed out; the sparsity mechanism does not eliminate the parameters of the whole layer to zero, so we consider the layer active.  
The sparsity pattern suggests that the layers of the backbone network capture the task-relevant features up to a certain depth, beyond which the performance of the model does not significantly benefit from the additional model complexity. 
Note that the sparsity patterns depend on the data (especially the amount of data) and the task.

\textit{Phase -2}: In phase 1, we identify the task-specific last non-zero layer, which we leverage in phase 2 to construct a Layer-optimized Multi-Task (LOMT) network.
Here, task-specific decoders are connected not at the end of the encoder but to their respective already identified intermediate layers, as depicted in Fig.~\ref{fig:block_diag}.
This hard parameter-sharing approach is motivated by the goal of harnessing efficient representations across multiple tasks.
Now that we have a \ac{LOMT} model, we train the network as it is done for any multi-task network \cite{paper2}, \cite{paper3}. 

Consider N non-identical but related tasks derived from a task distribution $p(T)$, say $\mathcal{T} = \{T_n\}_{n=1}^{N}$.
The objective of \ac{MTL} is to reduce the combined loss $\mathcal{L}_{comb}$, which is a function of the task-specific losses \ie $\{L_t\}_{t=1}^N$, and it can be represented as - $\min_{\theta} \mathcal{L}_{comb}(\theta),$ where, $\mathcal{L}_{comb}(\theta) = \mathcal{F}(\{L_t(\theta_{b},\theta_t)\}_{t=1}^N).$

Here, $\mathcal{F}$ represents the function or technique for loss balancing in a multi-task setting. 
There are several methods for loss balancing, which are discussed in \cite{crawshaw2020multi}.
In this work, we use the uncertainty-based loss balancing approach \cite{kendall2018multi}; according to this, the task-specific losses can be combined as, $\mathcal{L}_{comb} = \mathcal{F}({L_t}_{t=1}^{N}) = \sum_{t = 1}^{N} (L_t/2\sigma_t^2 + log~\sigma_t),$
where $\sigma_t$ is a learnable parameter for task $t$.
This work can, therefore, employ sparsity-induced feature selection for individual tasks, which is leveraged for multi-task learning. 
It should be emphasized that no parameters are transferred from phase 1 to phase 2. 
Only the layer information essential for building the Layer-Optimized Multi-Task (LOMT) models is obtained from phase 1.

\section{Experimental setup} \label{sec:exp_setup}

To validate the efficacy of the proposed work, we show the performance analysis on two publicly available \ac{MTL} datasets: the NYU-v2 dataset and CelebAMask-HQ dataset (mentioned as celebA dataset in this work). 
For the NYU-v2 dataset, we consider four dense prediction tasks: semantic segmentation($T_1$), depth estimation($T_2$), surface normal ($T_3$), and edge detection($T_4$).
For the celebA dataset, we consider one pixel-level task \ie semantic segmentation (3 classes - skin, hair, and background)($T_1$), and six image-level binary classification tasks that are male/no male($T_2$), smile/ no smile($T_3$), big-lips/no big-lips($T_4$), high cheekbones/ no high cheekbones($T_5$), wearing lipstick/ not wearing lipstick($T_6$), and bushy eyebrows/ no bushy eyebrows($T_7$).
Most of the works in the literature use all forty attributes in the celebA dataset as multiple tasks and report the average classification accuracy across all the tasks. 
However, we specifically chose some tasks focusing on different attributes to study the sparse feature selection; also, we reported the performance of all these tasks individually. 
For the celebA dataset, we consider combinations of all the tasks ($T_1 - T_7$), all classification tasks ($T_2 - T_7$), combinations of pixel level and image level tasks ($T_1,T_2,T_3,T_7 ~\text{and}~ T_1,T_4,T_7$), and some combinations of only classification tasks focusing on different attributes ($T_2,T_6,T_7 ~\text{and}~ T_2,T_6$).
For the NYU dataset, we have considered all the possible task combinations. 
For both the datasets, the network architecture, loss functions, and performance metrics are the same as used in other \ac{MTL} works such as \cite{sun2020adashare}, \cite{paper2}, \cite{paper3} for a fair comparison. 
As already mentioned in the previous section, we use dilated ResNet-50 as the backbone network. 
For dense prediction tasks, we have used a deeplab-v3 \cite{chen2017rethinking} network as the task-decoder or task-specific head, while for the classification tasks, a two-layer fully connected neural network is used as the task-specific network.

For a better performance comparison, we conducted four types of experiments: (i) dense \ac{STL}, (ii) sparse \ac{STL}, (iii) dense \ac{MTL}, (iv) \ac{LOMT} learning. 
Here, dense stands for the experiments without sparsity, considering all the layers of the backbone network. 
In the case of the NYU-v2 dataset, we have demonstrated the performance for all possible multi-task combinations, while for the celebA dataset, since the number of tasks is more (\ie 7), it was not feasible to show the performance for all the 120 multi-task combinations, so we present the results of a few of these combinations only. 
To ensure uniformity in assessments, all experiments conducted in this study employ identical architecture, loss functions, metrics, train-validation-test split, and hyperparameters.
All the experiments are conducted for five (random seed) runs, and we present the performance in terms of the mean and standard deviation values. 
The NVIDIA A100 Tensor Core GPUs, which have 40 GB of built-in HBM2 VRAM, are used to train the models.
For reproducibility, the source code is available at -- \href{https://github.com/ricupa/Layer-optimized-multi-task-model.git}{https://github.com/ricupa/Layer-optimized-multi-task-model.git}.


\section{Results and Discussion} \label{sec:discussion}
\begin{figure}[ht]
    \centering
    \includegraphics[width = \linewidth]{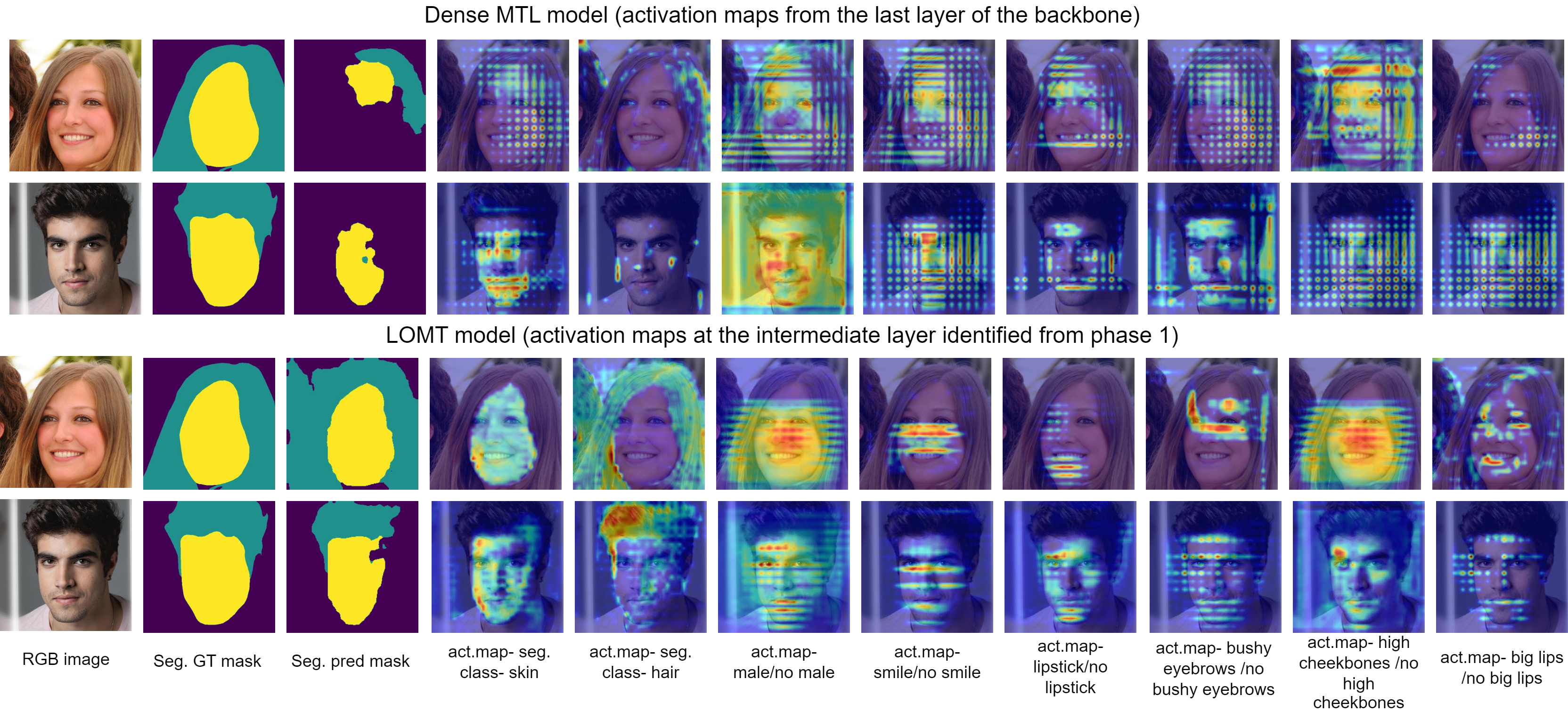}
    \caption{For the celebA dataset, the figure shows the RGB(input) image, the segmentation ground truth mask, and the predicted mask, along with the activation maps for the segmentation classes, as well as all the classification tasks for the dense \ac{MTL} and \ac{LOMT} models. Here act. map stands for the class activation maps\cite{jacobgilpytorchcam}.}
    \label{fig:grad_cam1}
    
\end{figure}

In this section, we present the results in the form of a comparative performance analysis of all the experiments mentioned in Sec.~\ref{sec:exp_setup} in Tables~\ref{tab:NYU_results_res} and \ref{tab:celebA_results}.
We present some qualitative results of the proposed approach; see Fig~\ref{fig:grad_cam1} and Table.~\ref{fig:NYU_all}.
The qualitative and quantitative performance and some ablation studies are discussed in this section hereafter. 

\noindent 
\textbf{Performance comparison with other methods}:
Most of the previous works have used three tasks in the case of the NYU-v2 dataset, so we only show the comparisons on those tasks.
Table.~\ref{tab:sota} shows that the proposed method outperforms the other few works on finding optimal multi-task architectures.

\noindent
\begin{minipage}[t]{0.55\textwidth}
\vspace{-1.2em}
\captionof{table}{Performance comparison of the proposed work with other works in the literature on the NYU-v2 dataset.}
\label{tab:sota}
\resizebox{\columnwidth}{!}{%
\begin{tabular}{l|c|c|ccc}
\hline
Models & \multicolumn{5}{c}{Tasks} \\ \hline
 & Segmen- & Depth & \multicolumn{3}{c}{Surface normal} \\ 
\textbf{} & tation & estimation &  \multicolumn{3}{c}{estimation}\\ 
 & & & &\multicolumn{2}{c}{angular error($\downarrow$}) \\ \cline{5-6}
\textbf{} & IoU($\uparrow$) & MAE($\downarrow$) & CS($\uparrow$ )& Mean & Median \\ \hline
Adashare \cite{sun2020adashare} & 30.2 & 0.55 & - & 16.6 & 12.9 \\
BMTAS \cite{bruggemann2020automated} & 41.1 & 0.54 & - & - & - \\
DAG-NAS \cite{choi2023dynamic} & 32.1 & 0.54 & - & 16.4 & 13.1 \\
MTL+sparsity\cite{paper3} & 33.94 & 0.13 & 0.78 & - & - \\ \hline
LOMT (ours) & \textbf{43.25} & \textbf{0.12} & \textbf{0.81} & \textbf{16.47} & \textbf{13.29} \\ \hline
\end{tabular}%
}
\rule{0pt}{8pt}
\end{minipage}%
\hfill 
\begin{minipage}[t]{0.43\textwidth} %

Table~\ref{tab:sota} presents a performance comparison with a selected few studies (discussed in Section~\ref{sec:related_work}) that have achieved results on the same datasets and tasks as ours. 
Note that the objective is not to compare the performance with the other \ac{MTL} methods in general; we emphasize those works that focus on feature selection for \ac{MTL}.
\end{minipage}


\begin{table}[]
\footnotesize
\centering
\caption{Performance of \ac{STL}, \ac{MTL}, and \ac{LOMT} experiments on the NYU-v2 dataset. Here IoU stands for intersection over union, MAE stands for mean absolute error, and CS stands for cosine similarity. The $\uparrow$ denotes that a higher value of the metric is preferable, while the $\downarrow$ denotes that a lower value is preferable.}
\label{tab:NYU_results_res}
\resizebox{0.75\columnwidth}{!}{%
\begin{tabular}{lclclclc}
\hline
\textbf{Experiments} & \multicolumn{7}{c}{\textbf{Tasks}} \\ \hline
\textbf{} & \textbf{Seg.} &  & \textbf{Depth} &  & \textbf{SN} &  & \textbf{Edge} \\
 & \textbf{$T_1$} &  & \textbf{$T_2$} &  & \textbf{$T_3$} &  & \textbf{$T_4$} \\
\textbf{} & IoU (↑) &  & MAE(↓) &  & CS (↑) &  & MAE(↓) \\ \hline
\multicolumn{8}{c}{\textbf{dense single-task model}} \\ \hline
\rule{0pt}{9pt}single task & 0.2814 $_{\pm 0.0032}$ &  & 0.1577 $_{\pm 0.0018}$ &  & 0.7241 $_{\pm 0.0013}$ &  & 0.5363 $_{\pm 0.0029}$ \\ \hline
\multicolumn{8}{c}{\textbf{sparse single-task model, $\lambda$ = 0.001}} \\ \hline
\rule{0pt}{9pt}single task & 0.2960 $_{\pm 0.0016}$ &  & 0.1555 $_{\pm 0.0011}$ &  & 0.7331 $_{\pm 0.0034}$ &  & 0.1807 $_{\pm 0.0090}$ \\
\begin{tabular}[c]{@{}c@{}}parameter \\ sparsity\end{tabular} & 66.3542 \% &  & 55.5215 \% &  & 41.7222 \% &  & 75.8073 \% \\ \hline
\multicolumn{8}{c}{\textbf{dense multi-task model}} \\ \hline
\rule{0pt}{9pt}$T_1$, $T_2$, $T_3$, $T_4$ & 0.2910 $_{\pm 0.0024}$ &  & 0.1502 $_{\pm 0.0006}$ &  & 0.7313 $_{\pm 0.0022}$ &  & 0.2168 $_{\pm 0.0020}$ \\
$T_1$,$T_2$,$T_3$ & 0.2914 $_{\pm 0.0041}$ &  & 0.1504 $_{\pm 0.0021}$ &  & 0.7325 $_{\pm 0.0031}$ &  &  \\
$T_2$,$T_3$,$T_4$ &  &  & 0.1423 $_{\pm 0.0062}$ &  & 0.7452 $_{\pm 0.0094}$ &  & 0.2152 $_{\pm 0.0098}$ \\ \hline
\multicolumn{8}{c}{\textbf{\rule{0pt}{7pt}LOMT model}} \\ \hline
\rule{0pt}{9pt}$T_1$, $T_2$, $T_3$, $T_4$ & \textbf{0.4272 $_{\pm 0.0022}$} &  & 0.1295 $_{\pm 0.0023}$ &  & 0.8077 $_{\pm 0.0015}$ &  & 0.1717 $_{\pm 0.0013}$ \\
$T_1$,$T_2$,$T_3$ & \textbf{0.4325 $_{\pm 0.0021}$} &  & \textbf{0.1248 $_{\pm 0.0008}$} &  & \textbf{0.8112 $_{\pm 0.0011}$} &  &  \\
$T_2$,$T_3$,$T_4$ &  &  & 0.1391 $_{\pm 0.0030}$ &  & 0.7986 $_{\pm 0.0018}$ &  & \textbf{0.1554 $_{\pm 0.0014}$} \\
$T_1$, $T_3$,$T_4$ & 0.4141 $_{\pm 0.0098}$ &  &  &  & \textbf{0.8111 $_{\pm 0.0046}$} &  & 0.1659 $_{\pm 0.0026}$ \\
$T_1$, $T_2$,$T_4$ & 0.4212 $_{\pm 0.0058}$ &  & \textbf{0.1292 $_{\pm 0.0023}$} &  &  &  & 0.1804 $_{\pm 0.0095}$ \\
$T_1$, $T_2$ & 0.4149 $_{\pm 0.0098}$ &  & 0.1304 $_{\pm 0.0029}$ &  &  &  &  \\
$T_1$, $T_3$ & 0.4035 $_{\pm 0.0078}$ &  &  &  & 0.8059 $_{\pm 0.0048}$ &  &  \\
$T_1$, $T_4$ & 0.3979 $_{\pm 0.0052}$ &  &  &  &  &  & 0.1649 $_{\pm 0.0015}$ \\
$T_2$, $T_3$ &  &  & 0.1422 $_{\pm 0.0012}$ &  & 0.7966 $_{\pm 0.0007}$ &  &  \\
$T_2$, $T_4$ &  &  & 0.1512 $_{\pm 0.0026}$ &  &  &  & \textbf{0.1519 $_{\pm 0.0004}$} \\
$T_3$, $T_4$ &  &  &  &  & 0.7992 $_{\pm 0.0030}$ &  & 0.1711 $_{\pm 0.0019}$ \\ \hline
\end{tabular}%
}

\end{table}
\begin{table}[]
\footnotesize
\centering
\caption{Performance of \ac{STL}, \ac{MTL}, and \ac{LOMT} experiments on the celebA dataset. Here IoU stands for intersection over union, and acc. stands for binary classification accuracy. Both both the metrics, a higher value (denoted by $\uparrow$) is preferable.}
\label{tab:celebA_results}
\resizebox{\columnwidth}{!}{%
\begin{tabular}{lccccccc}
\hline
\textbf{\begin{tabular}[c]{@{}c@{}}Experi-\\ ments\end{tabular}} & \textbf{\begin{tabular}[c]{@{}c@{}}Segmen-\\ tation\end{tabular}} & \textbf{male} & \textbf{smile} & \textbf{biglips} & \textbf{\begin{tabular}[c]{@{}c@{}}high\\ cheekbones\end{tabular}} & \textbf{\begin{tabular}[c]{@{}c@{}}wearing\\ lipstick\end{tabular}} & \textbf{\begin{tabular}[c]{@{}c@{}}bushy\\ eyebrows\end{tabular}} \\
& \textbf{(3 class)} & \textbf{(0/1)} & \textbf{(0/1)} & \textbf{(0/1)} & \textbf{(0/1)} & \textbf{(0/1)} & \textbf{(0/1)} \\
 & $T_1$ & $T_2$ & $T_3$ & $T_4$ & $T_5$ & $T_6$ & $T_7$ \\
 & IoU($\uparrow$) & acc.($\uparrow$) & acc.($\uparrow$) & acc.($\uparrow$) & acc.($\uparrow$) & acc.($\uparrow$) & acc.($\uparrow$) \\ \hline
\multicolumn{8}{c}{\textbf{dense single-task model}} \\ \hline
\rule{0pt}{10pt}STL & \textbf{0.9195 $_{\pm0.0001}$} & 0.9648 $_{\pm0.0044}$ & \textbf{0.8987 $_{\pm0.0016}$} & \textbf{0.6358 $_{\pm0.0061}$} & 0.8273 $_{\pm0.0101}$ & 0.9106 $_{\pm0.0049}$ & 0.6270 $_{\pm0.0100}$ \\ \hline
\multicolumn{8}{c}{\textbf{sparse single-task model, $\lambda$ = 0.001}} \\ \hline
\rule{0pt}{10pt}STL & 0.8685 $_{\pm0.0033}$ & 0.7773 $_{\pm0.0971}$ & 0.8569 $_{\pm0.0127}$ & \textbf{0.6259 $_{\pm0.0069}$} & 0.5433 $_{\pm0.0167}$ & 0.7502 $_{\pm0.0486}$ & 0.6212 $_{\pm0.0023}$ \\
\begin{tabular}[c]{@{}c@{}}parameter \\ sparsity\end{tabular} & 82.27\% & 81.55\% & 78.71\% & 71.61\% & 81.81\% & 83.22\% & 74.28\% \\\hline
\multicolumn{8}{c}{\textbf{dense multi-task model}} \\ \hline
\rule{0pt}{10pt}$T_1$-$T_7$ & 0.8941 $_{\pm0.0027}$ & 0.9554 $_{\pm0.0055}$ & 0.8909 $_{\pm0.0047}$ & 0.6340 $_{\pm0.0003}$ & \textbf{0.8454 $_{\pm0.0010}$} & 0.9122 $_{\pm0.0031}$ & 0.7867 $_{\pm0.0016}$ \\
$T_2$-$T_7$ & - & 0.9496 $_{\pm0.0048}$ & 0.8790 $_{\pm0.0128}$ & 0.6337 $_{\pm0.0000}$ & 0.8327 $_{\pm0.0217}$ & 0.9034 $_{\pm0.0084}$ & 0.7814 $_{\pm0.0154}$ \\
$T_1$,$T_2$,$T_3$,$T_7$ & 0.8995 $_{\pm0.0014}$ & 0.9513 $_{\pm0.0010}$ & 0.8901 $_{\pm0.0017}$ & - & - & - & 0.7890 $_{\pm0.0087}$ \\ \hline
\multicolumn{8}{c}{\textbf{LOMT models}} \\ \hline
\rule{0pt}{10pt}$T_1$-$T_7$ & 0.9123 $_{\pm0.0005}$ & 0.9681 $_{\pm0.0034}$ & \textbf{0.8967 $_{\pm0.0020}$} & 0.6345 $_{\pm0.0020}$ & \textbf{0.8404 $_{\pm0.0029}$} & \textbf{0.9135 $_{\pm0.0042}$} & 0.7985 $_{\pm0.0067}$ \\
$T_2$-$T_7$ & - & 0.9597 $_{\pm0.0019}$ & 0.8766 $_{\pm0.0082}$ & 0.6345 $_{\pm0.0056}$ & 0.8247 $_{\pm0.0137}$ & 0.9056 $_{\pm0.0019}$ & 0.7846 $_{\pm0.0165}$ \\
$T_1$,$T_2$,$T_3$,$T_7$ & \textbf{0.9136 $_{\pm0.0005}$} & 0.9683 $_{\pm0.0034}$ & 0.8942 $_{\pm0.0080}$ & - & - & - & \textbf{0.8068 $_{\pm0.0090}$} \\
$T_2$,$T_6$,$T_7$ & - & \textbf{0.9685 $_{\pm0.0044}$} & 0.8850 $_{\pm0.0064}$ & - & - & 0.9096 $_{\pm0.0022}$ & - \\
$T_2$,$T_6$ & - & \textbf{0.9688 $_{\pm0.0042}$} & - & - & - & \textbf{0.9169 $_{\pm0.0050}$} & - \\
$T_1$,$T_4$,$T_7$ & 0.8897 $_{\pm0.0029}$ & - & - & 0.6300 $_{\pm0.0028}$ & - & - & \textbf{0.8034 $_{\pm0.0225}$} \\ \hline
\end{tabular}%
}

\end{table}

\textbf{\ac{MTL} vs \ac{STL}}: The performance comparisons between dense \ac{STL} and \ac{MTL} in Tables~\ref{tab:NYU_results_res} and \ref{tab:celebA_results} reveal that \ac{MTL} is equally effective or, in some cases, outperforms \ac{STL} across various tasks.
This is because of the inductive transfer property of \ac{MTL}, where the tasks share information to improve the learning of other related tasks. 
Due to its ability to leverage shared information to improve performance, \ac{MTL} emerges as one of the favorable approaches for achieving resource-efficient architectures and potentially incomparable learning outcomes in complex multi-task settings. 
However, it is crucial to acknowledge that some tasks may exhibit suboptimal performance when trained in a multi-task setting due to negative information transfer, where related tasks inadvertently compete with each other, resulting in a decline in their performance.

\begin{table}{}
\centering
\caption{Comparison of the qualitative performance of the single task dense (Dense STL), multi-task dense (Dense MTL) and the proposed LOMT model (Ours) for four tasks of the NYU-v2 dataset}
\label{fig:NYU_all}
\begin{tabular}{cccccc}
 &\textbf{Input} &\textbf{Ground} & \textbf{Dense} & \textbf{Dense } & \textbf{LOMT} \\
  &\textbf{image} & \textbf{truth} & \textbf{STL} & \textbf{MTL} & \textbf{(Ours)} \\
\rotatebox{90}{\textbf{Seg.($T_1$)}} & \includegraphics[width=0.13\textwidth]{figures/NYU/image.jpg} & \includegraphics[width=0.13\textwidth]{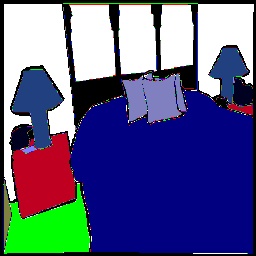} & \includegraphics[width=0.13\textwidth]{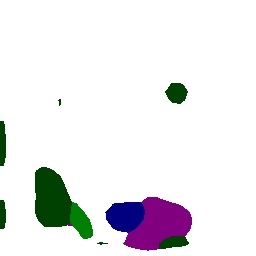} & \includegraphics[width=0.13\textwidth]{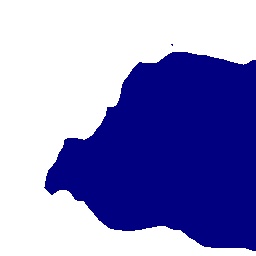} & \includegraphics[width=0.13\textwidth]{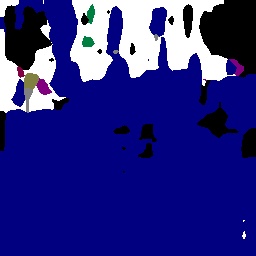} \\
\rotatebox{90}{\textbf{Depth($T_2$)}}& \includegraphics[width=0.13\textwidth]{figures/NYU/image.jpg} & \includegraphics[width=0.13\textwidth]{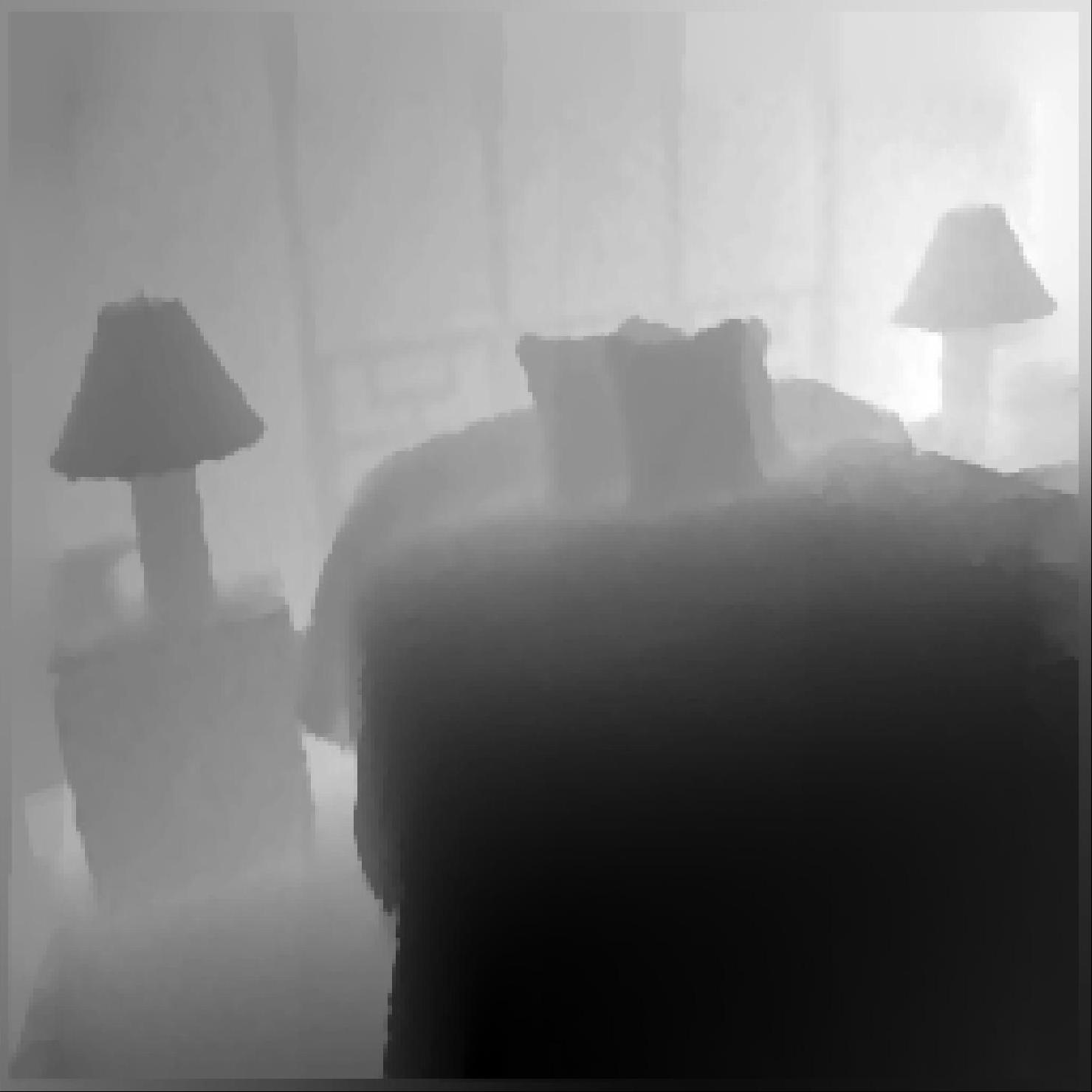} & \includegraphics[width=0.13\textwidth]{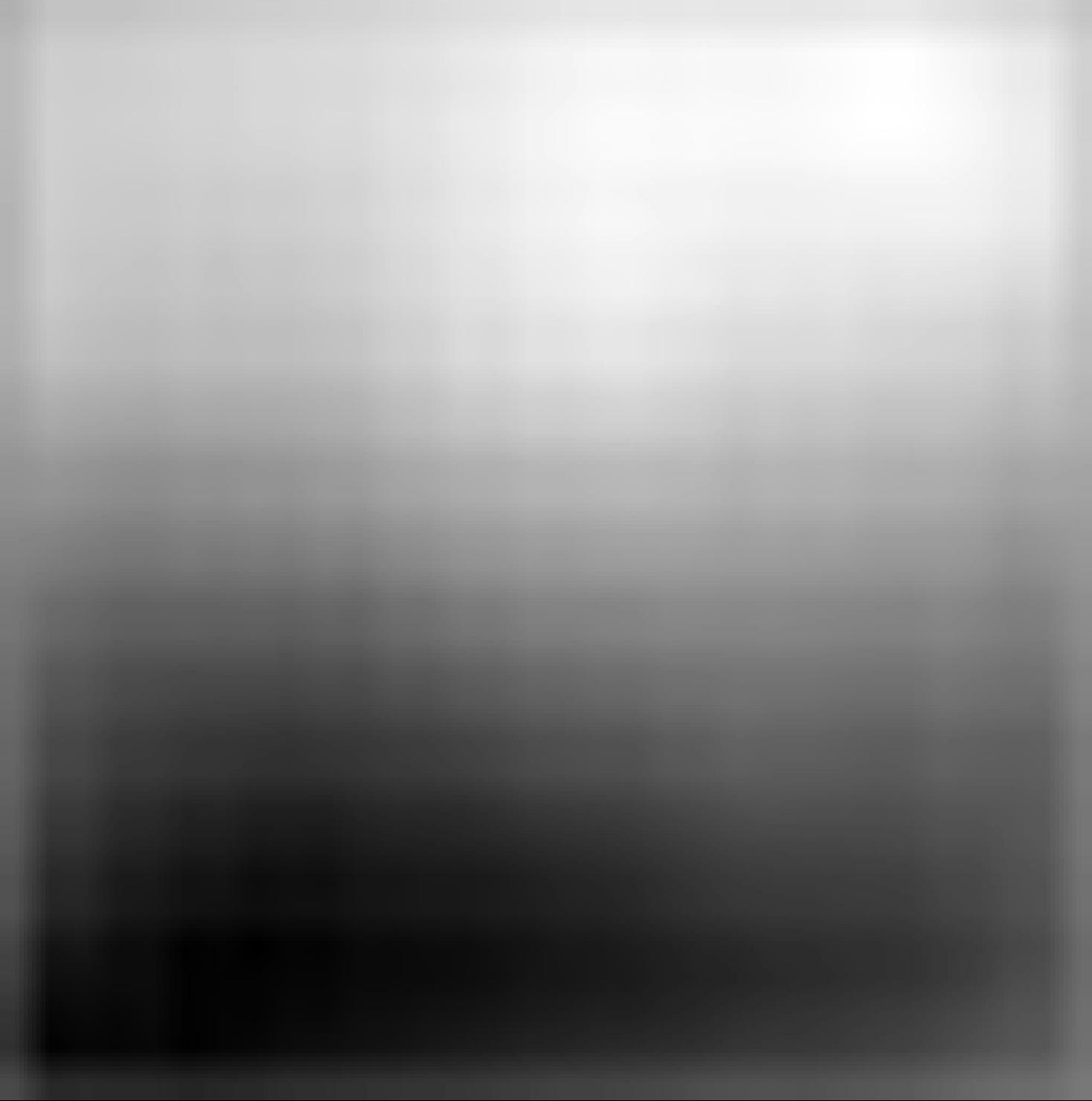} & \includegraphics[width=0.13\textwidth]{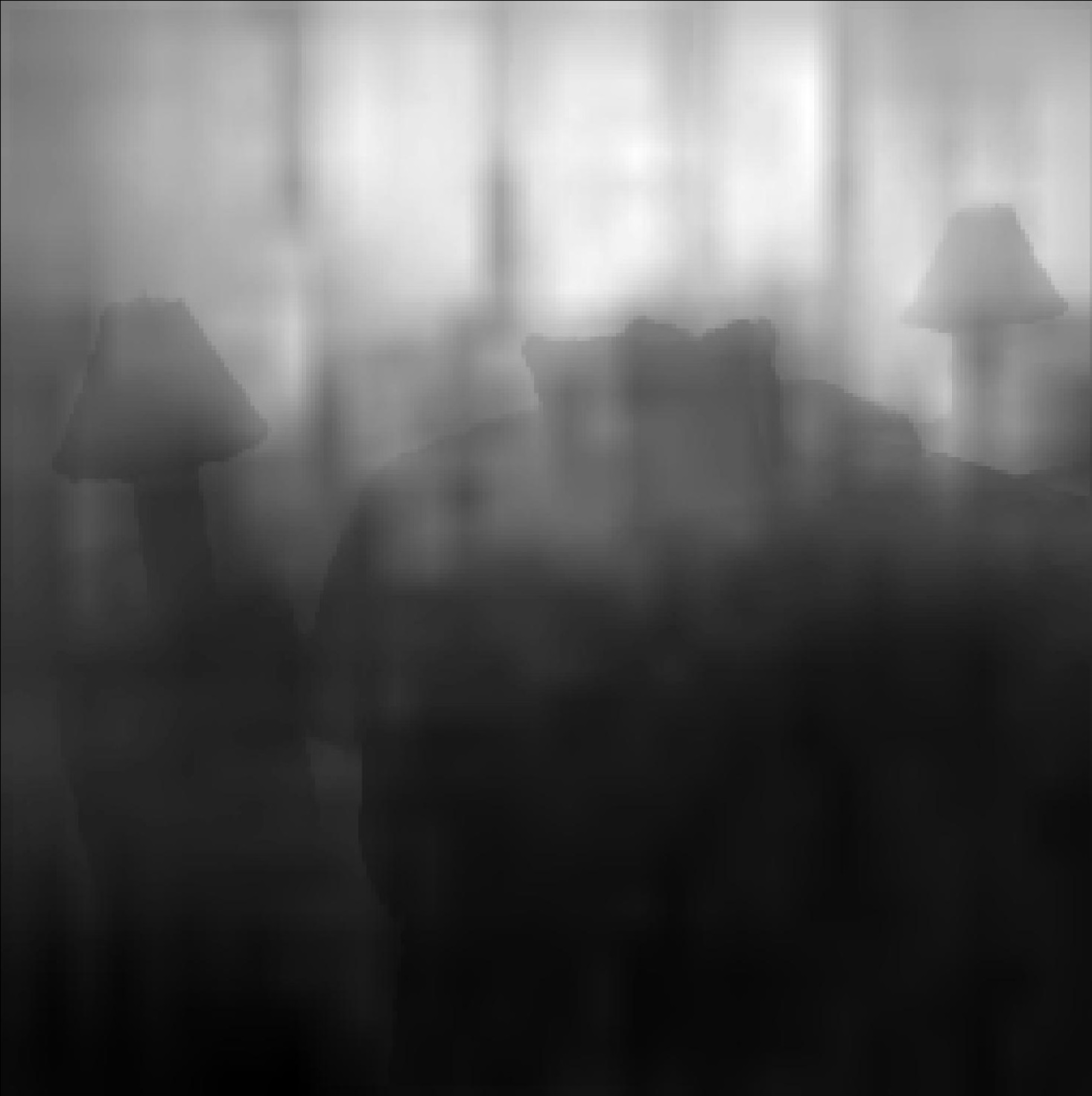} & \includegraphics[width=0.13\textwidth]{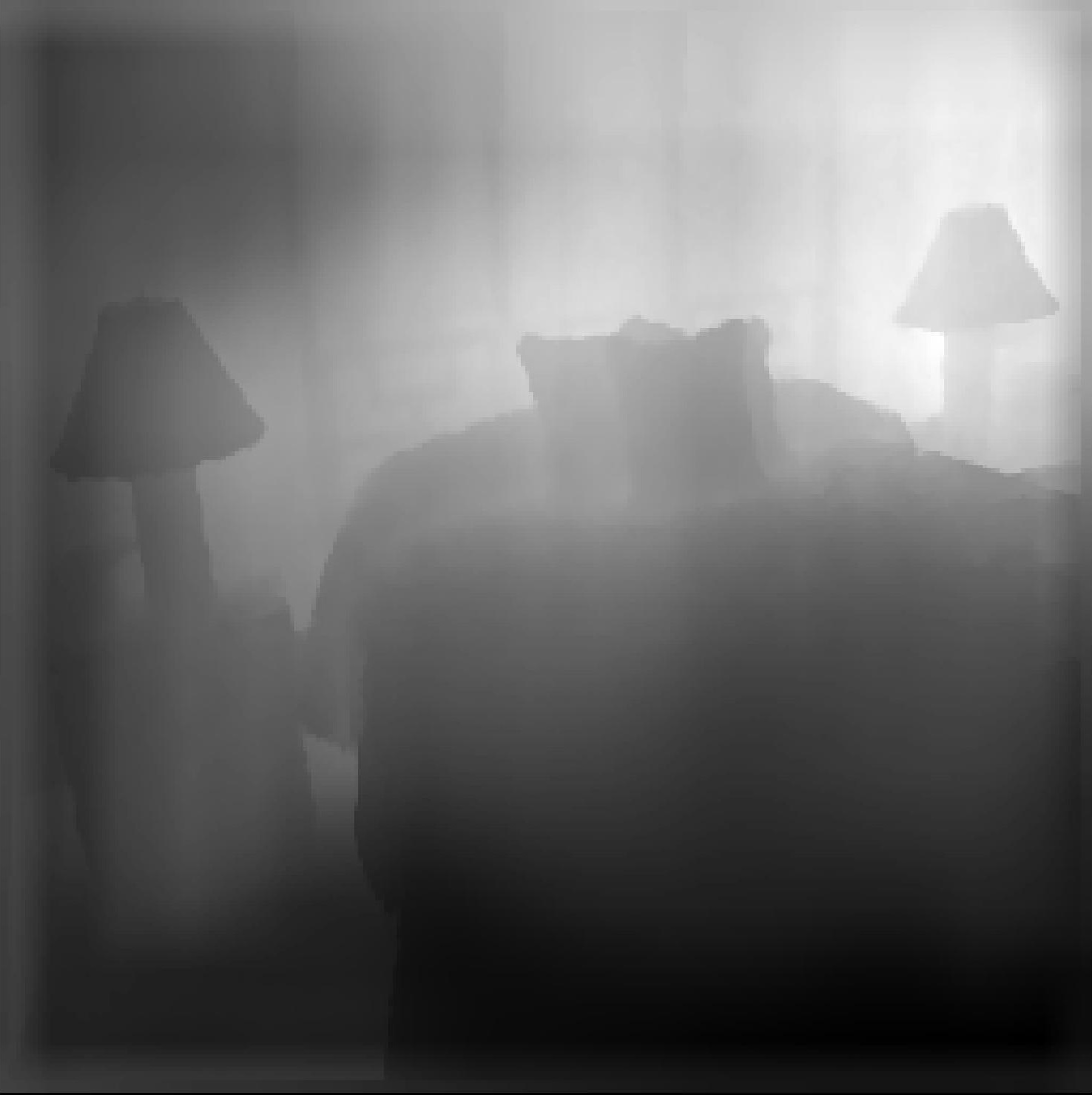} \\
\rotatebox{90}{\textbf{SN($T_3$)}}& \includegraphics[width=0.13\textwidth]{figures/NYU/image.jpg} & \includegraphics[width=0.13\textwidth]{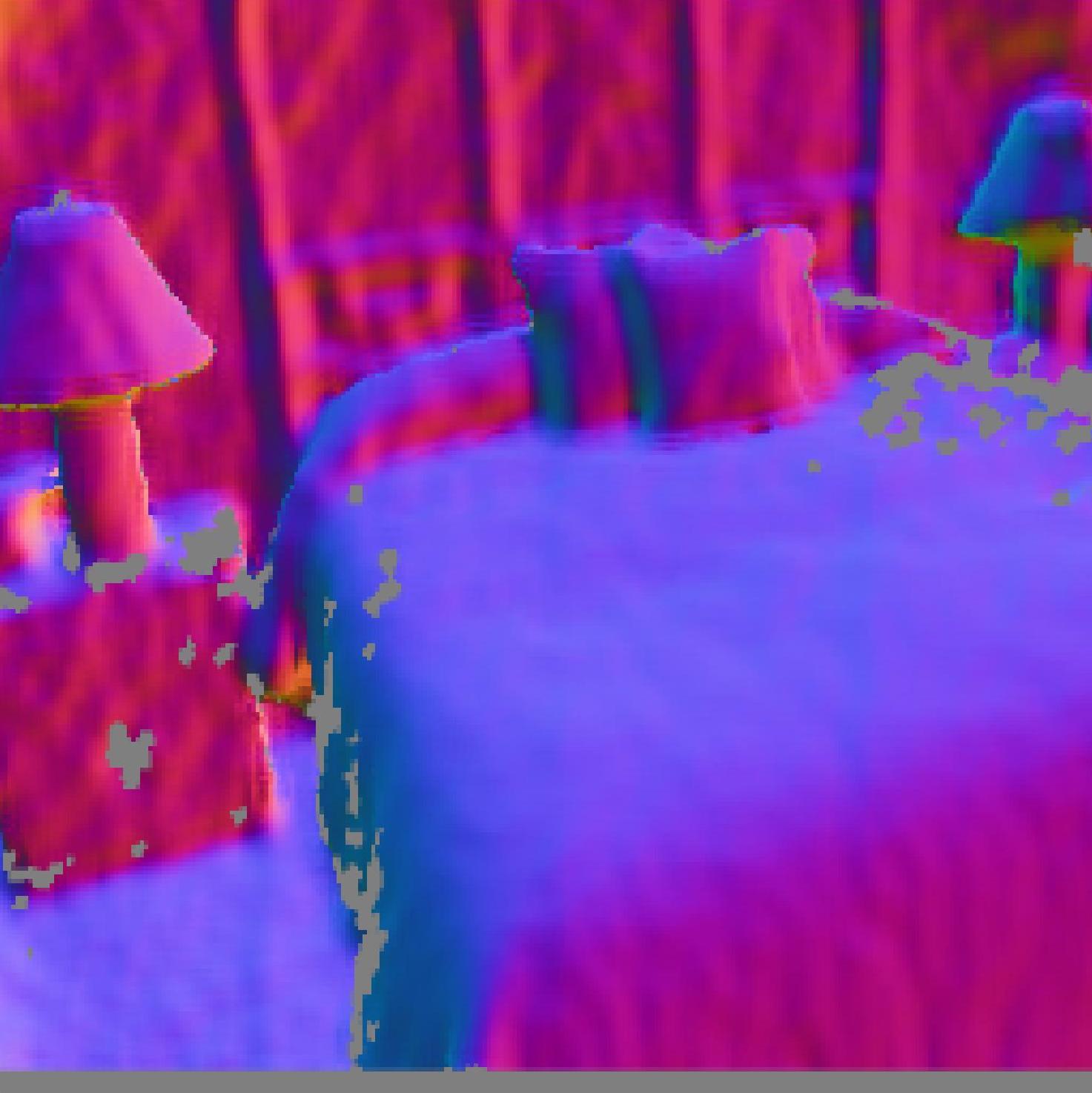} & \includegraphics[width=0.13\textwidth]{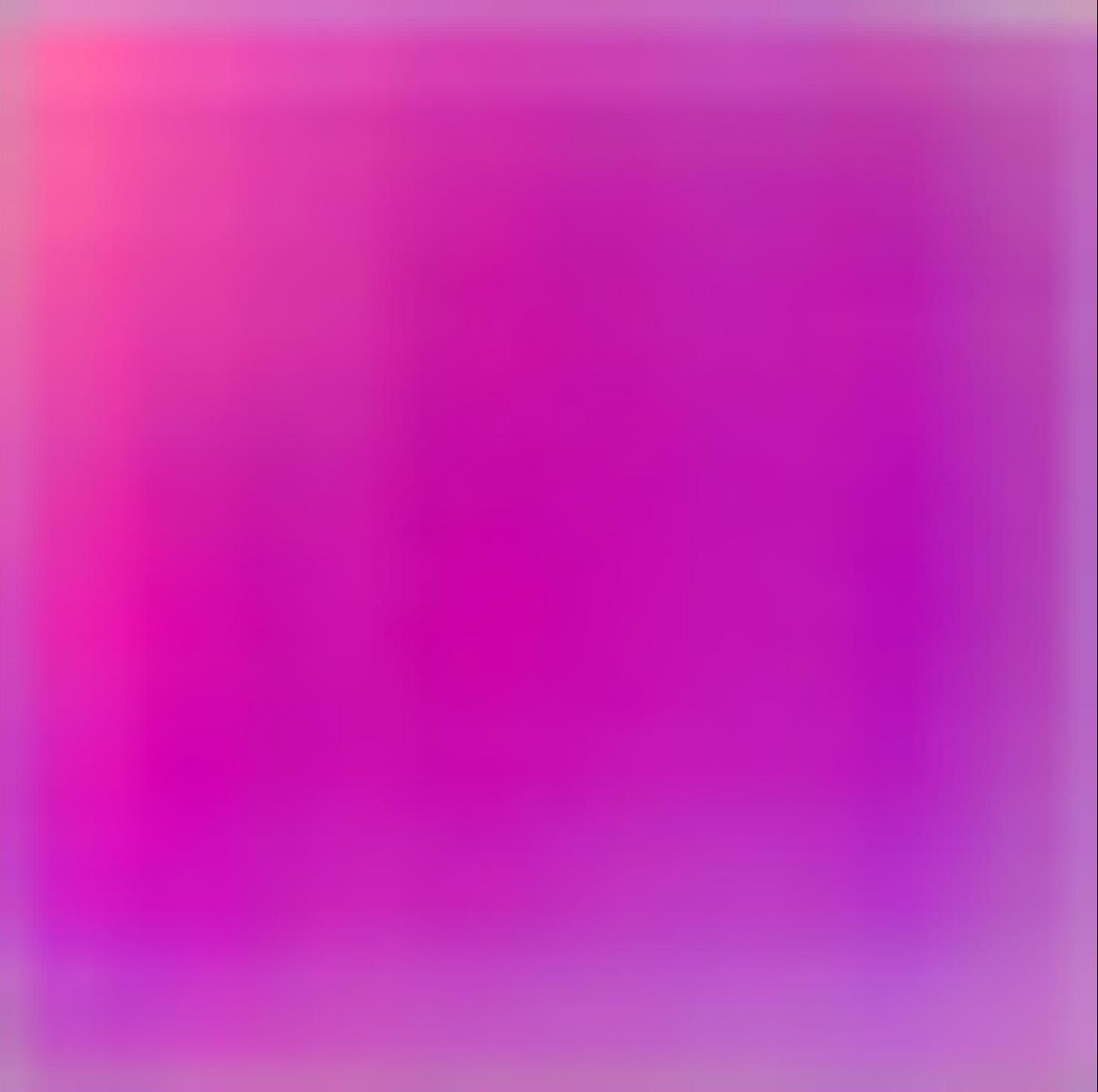} & \includegraphics[width=0.13\textwidth]{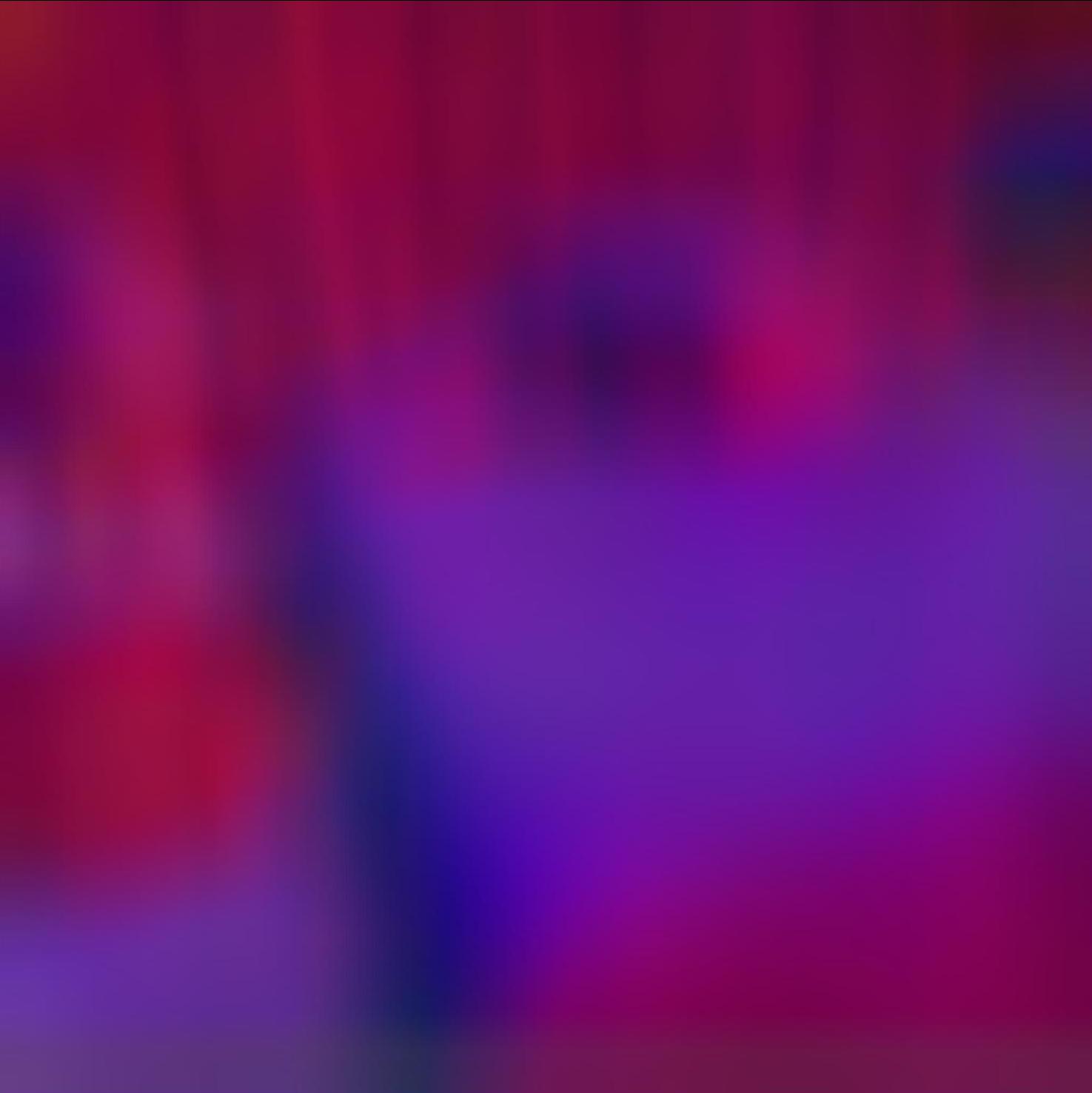} & \includegraphics[width=0.13\textwidth]{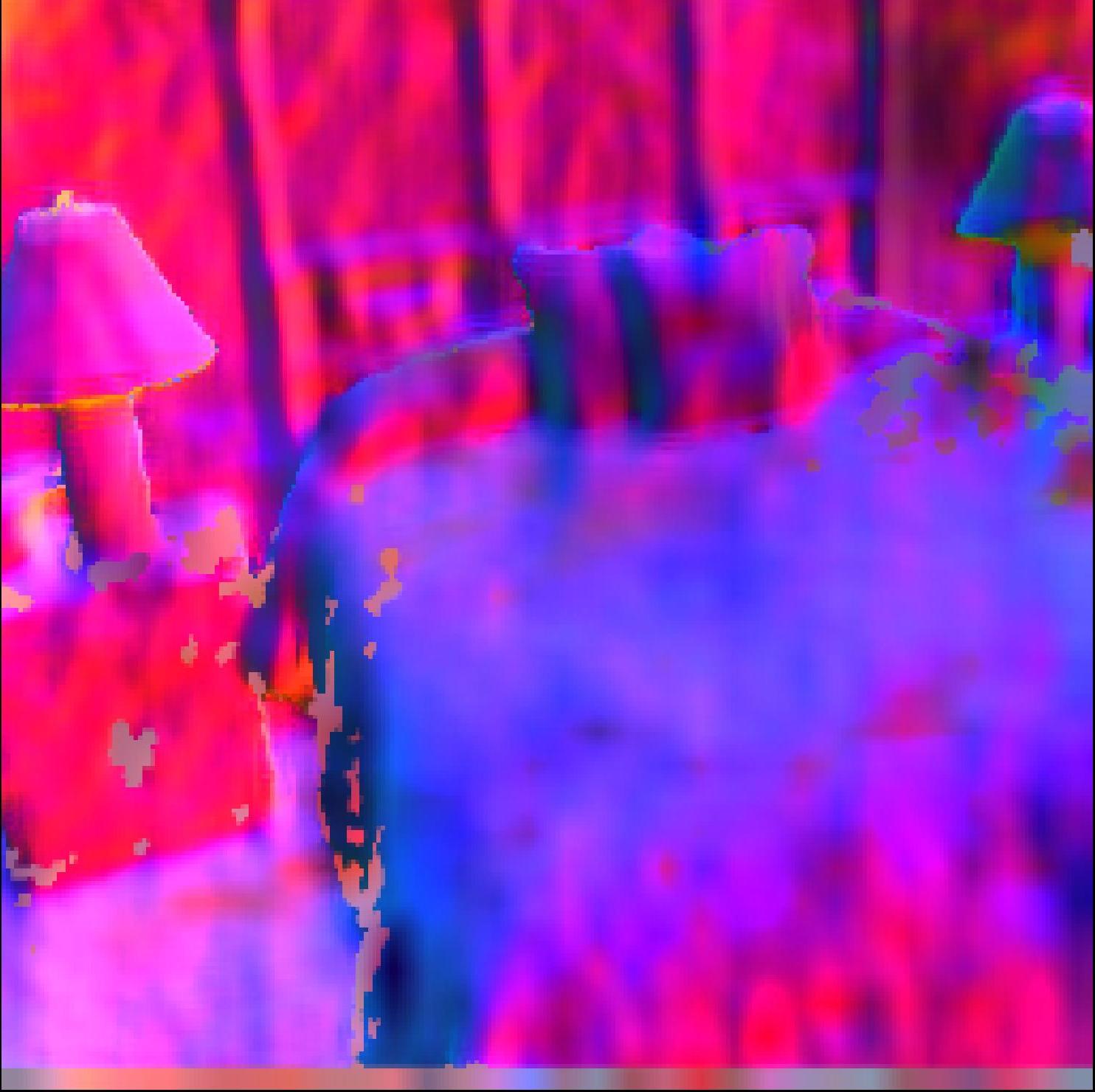} \\
\rotatebox{90}{\textbf{Edge($T_4$)}}& \includegraphics[width=0.13\textwidth]{figures/NYU/image.jpg} & \includegraphics[width=0.13\textwidth]{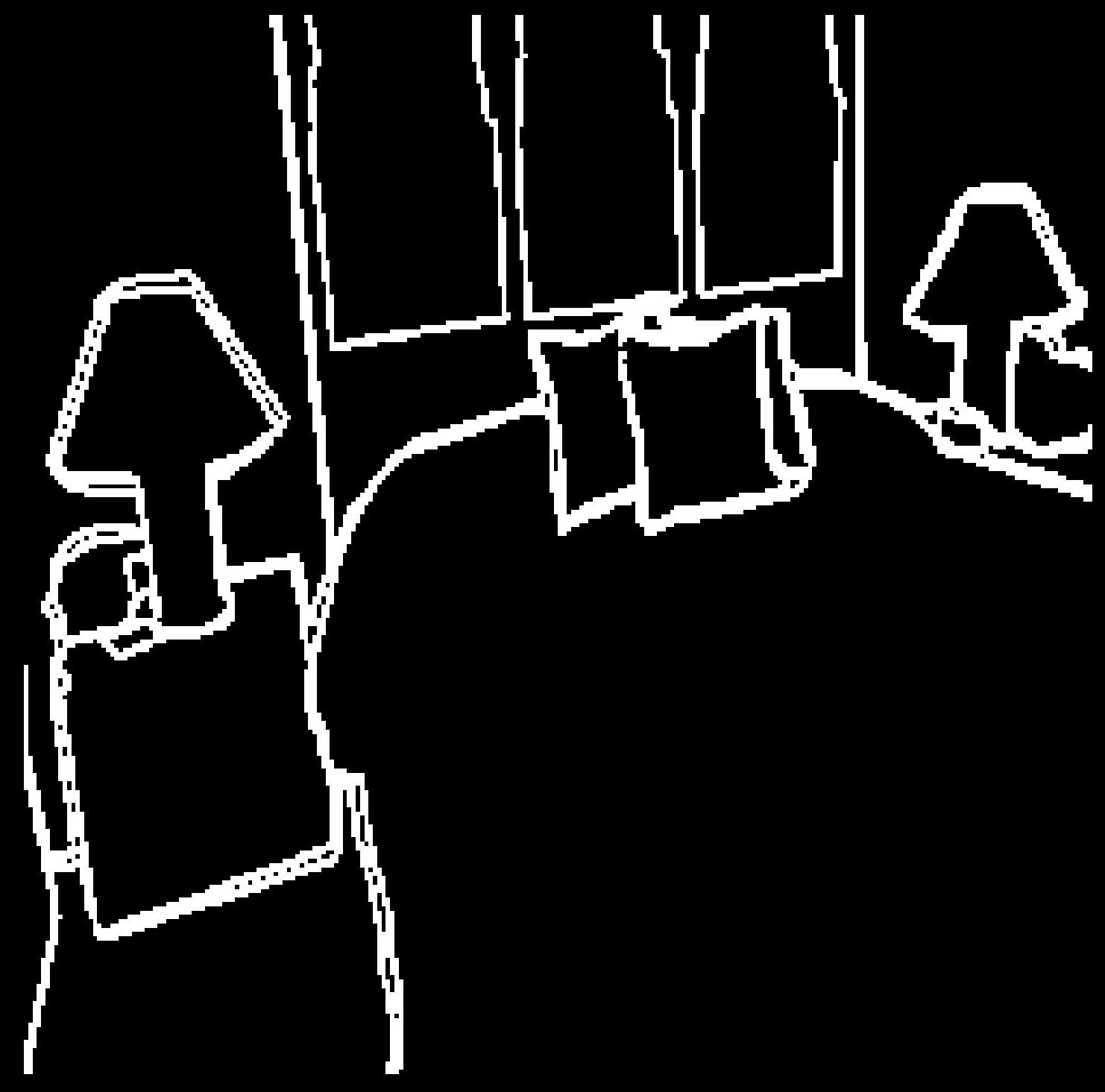} & \includegraphics[width=0.13\textwidth]{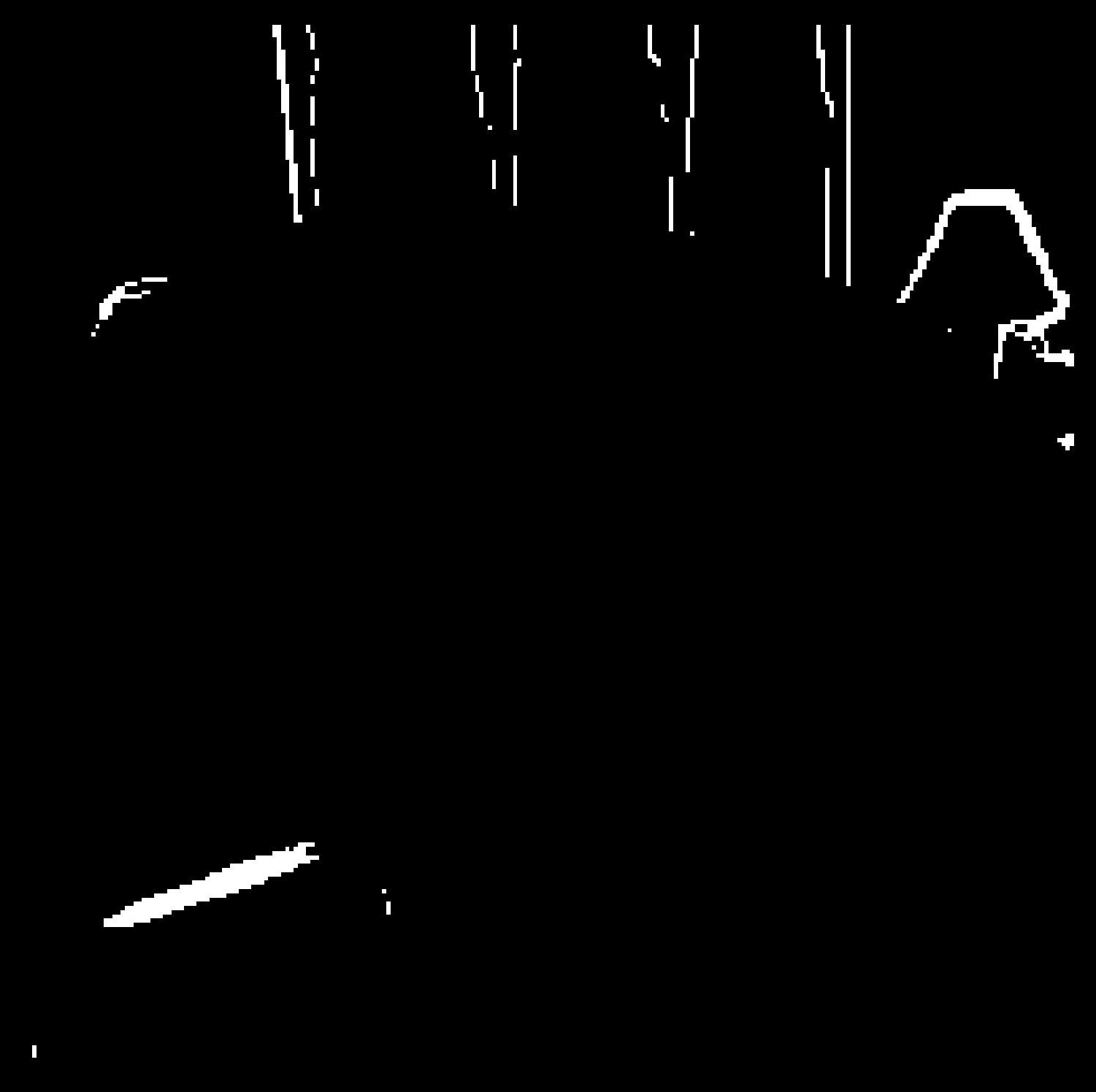} & \includegraphics[width=0.13\textwidth]{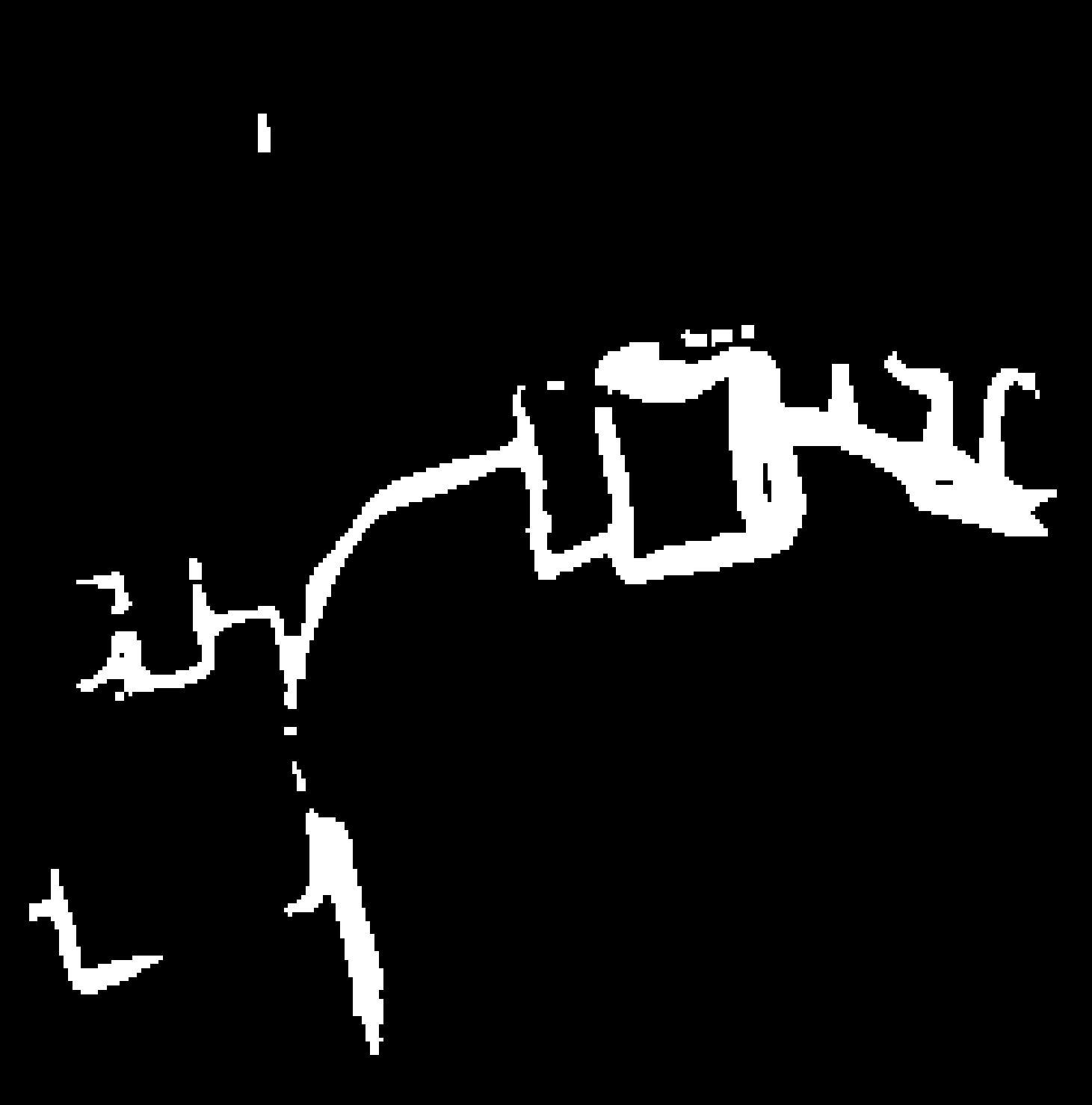} & \includegraphics[width=0.13\textwidth]{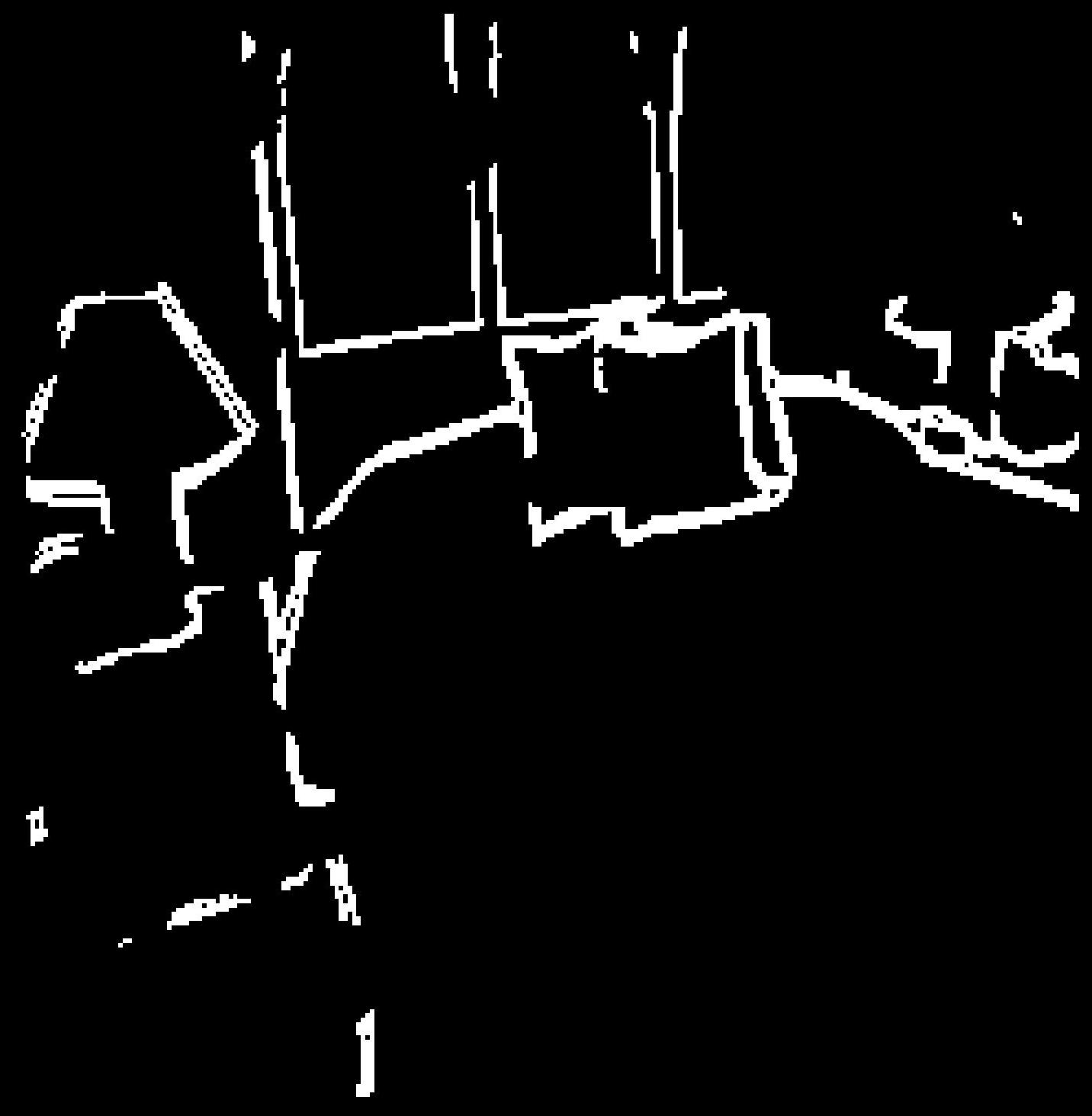} \\
\end{tabular}

\end{table}

\textbf{Effect of group sparsity on \ac{STL}}: 
For the NYU-v2 dataset \ie Table~\ref{tab:NYU_results_res}, when group sparsity is applied in the encoder network, a significant proportion of the parameters are set to zero, represented by `parameter sparsity\footnote[1]{In this work, parameter sparsity or \% parameter sparsity stands for the proportion of the parameters set to zero.}' in the table. 
Additionally, the performance of the tasks is improved as compared to dense \ac{STL}. 
Fig.~\ref{fig:pattern} illustrates the sparsity pattern of the backbone network for all four tasks. 
It is understood from this figure that all tasks have different feature requirements.
For example, edge detection, which involves identifying discontinuities in an image corresponding to object boundaries or texture changes, relies on the low-level features of the network that are extracted by the lower layers of the network. 
Segmentation requires mid-level features since it requires not only edge features and object boundaries but also textures, colors, and shapes.
Surface normal detection requires more complex features learned at deeper layers in a network as compared to segmentation because it involves interpreting the scene's geometry.
Depth estimation requires a nuanced understanding of the image that captures spatial relationships and object shapes, which is more than edge detection but less than segmentation or surface normal estimation. 
This analysis is the prime motivation behind this work. 

For the celebA dataset \ie Table.~\ref{tab:celebA_results}, the performance of experiments with sparsity is suboptimal for many tasks compared to \ac{STL}. 
This can be attributed to the backbone's reduced complexity caused by sparsity, which limits its ability to capture the vital features required for these tasks.
This indicates that not all tasks require the same level or strength of sparsity (in terms of the value of $\lambda$).
Nevertheless, we choose to proceed with phase 2 of this approach to validate the efficacy of the LOMT models. 
\textbf{Performance of \ac{LOMT}}:
In phase 2 of the proposed approach, we design the multi-task model \ie \ac{LOMT} models using the optimal layer information from phase 1 and connect the task heads or decoders to these intermediate layers.
This technique tailors the network architecture to the specific requirements of each task, hence improving the learning process and the performance of the tasks. 
This is evident from the results in Tables~\ref{tab:NYU_results_res} and \ref{tab:celebA_results}. 
For all the task combinations, we observe that LOMT models outperform their dense \ac{STL} and \ac{MTL} counterparts, especially for the NYU-v2 dataset. 
We observe that some task combinations work better than others primarily because of their conflicting requirements.
For example, semantic segmentation ($T_1$) and edge detection ($T_4$) do not perform their best together.
$T_4$ outperforms in combinations that do not have $T_1$, similarly the best IoU for $T_1$ is in the ensemble without $T_4$. 
Overall, the remarkable improvements in the performance metrics indicate that \ac{LOMT} models benefit from the targeted layer selection and can utilize the features specific to each task. 
This claim is also supported by the qualitative performance of the \ac{LOMT} models; see Table~\ref{fig:NYU_all}.
For all the tasks, the outputs of \ac{LOMT} models are much better than the others. 
The segmentation output is not up to the mark for any other of the models, which is seconded by the low value of the IoU metric; this is probably because the segmentation masks for this dataset are very dense, having 40 segmentation classes. 
\\
\indent For the celebA dataset, the performance of \ac{LOMT} models is comparable or, for some tasks, better when compared to the dense and sparse models. 
Although the performance of sparse \ac{STL} falls below that of their dense counterparts, underscoring the challenge of maintaining performance under high sparsity conditions.
However, \ac{LOMT} models show comparable performance to their dense counterparts, indicating the importance of training multiple tasks jointly and, hence, striking a balance between model complexity reduction and task-specific feature learning.
\ac{LOMT} models show noticeable but not extreme improvements in performance (as for the NYU-v2 dataset).
This performance trend can be due to negative information transfer across tasks.
Due to the distinct nature of the tasks, such as segmentation being a pixel-level task while others are image-level tasks, and the diverse focus of the binary-classification tasks (\eg male/no-male considering the entire image versus high cheekbones, smile, etc., focusing on specific regions of the face), there may be conflicting requirements that hinder the learning in the multi-task setting.
\\
\indent Fig.~\ref{fig:grad_cam1} compares the qualitative results of \ac{LOMT} models to the dense \ac{MTL} models.
It shows the predicted segmentation masks for the sample input RGB image and the activation maps for segmentation classes and binary classification tasks. 
Activation maps give insight into which parts of the input image the model focuses on for a particular prediction \cite{jung2021towards}.
We use GradCam \cite{8237336}, \cite{jacobgilpytorchcam} to visualize the activation maps.
The activation maps are determined from the last layer of the backbone network for the dense \ac{MTL}, while for the \ac{LOMT} models, they are calculated at the layer from where the respective task head(encoder) is connected.
The activation maps in Fig.~\ref{fig:grad_cam1} demonstrate that for the \ac{LOMT} models, the activations are very focused according to the task.
In the case of the dense \ac{MTL}, they are scattered across the image for most of the tasks, which may imply that the model is considering a wide range of features across the entire image for its prediction rather than focusing on a few distinct areas.
\begin{table}[t]
\centering
\caption{Parameter counts and Compression Ratio(CR) for various tasks and their combinations for STL, MTL, and LOMT models for the NYU dataset.}
\label{tab:CRtable}
\resizebox{\textwidth}{!}{%
\begin{tabular}{lccccccc}
\toprule
\multicolumn{8}{c}{\textbf{Backbone (encoder) parameters only}}                                                                                                                                    \\ \toprule
\textbf{Tasks}         &               & \multicolumn{2}{c}{\textbf{\#parameters in Million}} & & \multicolumn{3}{c}{\textbf{CR = \#total parameters/\#nonzero pramaters}}                         \\ \cmidrule{1-1} \cmidrule{3-4} \cmidrule{6-8}
\textbf{}           &                  & \textbf{dense}            & \textbf{sparse}           &  & \textbf{Compression Ratio (CR)} & \multicolumn{1}{l}{\textbf{}} & \multicolumn{1}{l}{}             \\
\textbf{}           &                  & \textbf{STL}              & \textbf{STL}       &       & \textbf{dense vs sparse STL}    & \multicolumn{1}{l}{\textbf{}} & \multicolumn{1}{l}{}             \\ \hline
$T_1$              &                      & 23.63                     & 7.29                  &    & 3.24                            & \multicolumn{1}{l}{}          & \multicolumn{1}{l}{}             \\
$T_2$              &                      & 23.63                     & 5.58                  &    & 4.23                            & \multicolumn{1}{l}{}          & \multicolumn{1}{l}{}             \\
$T_3$               &                     & 23.63                     & 9.19                  &    & 2.57                            & \multicolumn{1}{l}{}          & \multicolumn{1}{l}{}             \\
$T_4$               &                     & 23.63                     & 1.5                   &    & 15.75                           & \multicolumn{1}{l}{}          & \multicolumn{1}{l}{}             \\ \hline
\textbf{}         &               & \textbf{}            & \textbf{}          &   & \multicolumn{3}{c}{\textbf{Compression Ratio (CR)}}                                                \\ \cmidrule{6-8} 
\textbf{Tasks}      &                 & \textbf{dense}              & \textbf{LOMT}     & \textbf{}            & \textbf{dense-MTL}              & \textbf{dense-STL$\times$\#tasks}  & \textbf{dense-STL$\times$\#tasks}     \\
\textbf{}      &                & \textbf{MTL}                 & \textbf{}      & \textbf{}           & \textbf{vs LOMT}                & \textbf{vs LOMT}              & \textbf{vs spars-STL of tasks} \\ \hline
{\color[HTML]{444444} $T_1$, $T_2$, $T_3$, $T_4$}&  & 23.63                     & 9.19          &            & 2.57                            & 10.29                         & 4.01                             \\
$T_1$,$T_2$,$T_3$          &                     & 23.63                     & 9.19             &         & 2.57                            & 7.71                          & 4.59                             \\
$T_2$,$T_3$,$T_4$        &                       & 23.63                     & 9.19            &          & 2.57                            & 7.71                          & 2.32                             \\
$T_1$, $T_3$,$T_4$       &                       & 23.63                     & 9.19             &         & 2.57                            & 7.71                          & 3.94                             \\
$T_1$, $T_2$,$T_4$        &                      & 23.63                     & 7.29            &          & 3.24                            & 9.72                          & 4.93                             \\
$T_1$, $T_2$            &                     & 23.63                     & 7.29              &        & 3.24                            & 6.48                          & 3.67                             \\
$T_1$, $T_3$           &                      & 23.63                     & 7.29             &         & 3.24                            & 6.48                          & 2.87                             \\
$T_1$, $T_4$            &                     & 23.63                     & 7.29             &         & 3.24                            & 6.48                          & 5.38                             \\
$T_2$, $T_3$            &                     & 23.63                     & 9.19            &          & 2.57                            & 5.14                          & 3.20                             \\
$T_2$, $T_4$             &                    & 23.63                     & 5.58            &          & 4.23                            & 8.47                          & 6.68                             \\
$T_3$, $T_4$              &                   & 23.63                     & 9.19             &         & 2.57                            & 5.14                          & 4.42                             \\ 
\bottomrule
\end{tabular}%
}
\vspace{-2em}
\end{table}

\textbf{Model compression}: 
Table~\ref{tab:CRtable} provides insights on the total number of parameters in a dense backbone, the number of non-zero parameters in the backbone of a sparse single task model,s, and that of the proposed \ac{LOMT} models. 
Given that the same decoders or task-specific heads are used across all the experiments for all tasks, we solely evaluate and compare the backbone (encoder) parameters that incorporate structural sparsity. 
A widely employed metric for assessing compression in deep learning models is the \textit{compression ratio} (CR), defined as the total number of parameters divided by the number of non-zero parameters \cite{blalock2020state}. A larger value indicates better compression.
In Table~\ref{tab:CRtable}, we assess the degree of compression resulting from the \ac{LOMT} models on the dense multi-task and single-task models. 
We also give the compression when sparse single-task models are used instead of their dense counterparts. 
The impact of LOMT models on compressing dense MTL models is minimal; however, when it comes to dense STL models, the compression is substantial. 
The model compression resulting from MTL is far larger than that from sparsification.
This study uses the combination of \ac{MTL} and structured sparsity. 
It is evident that \ac{MTL} mainly contributed to model compression, while sparsity played a role in selecting task-specific features. This feature selection ultimately improved the performance of all tasks in a multi-task context. 

\begin{figure}[t]
  \begin{minipage}{0.43\textwidth}
    \includegraphics[width=0.85\linewidth]{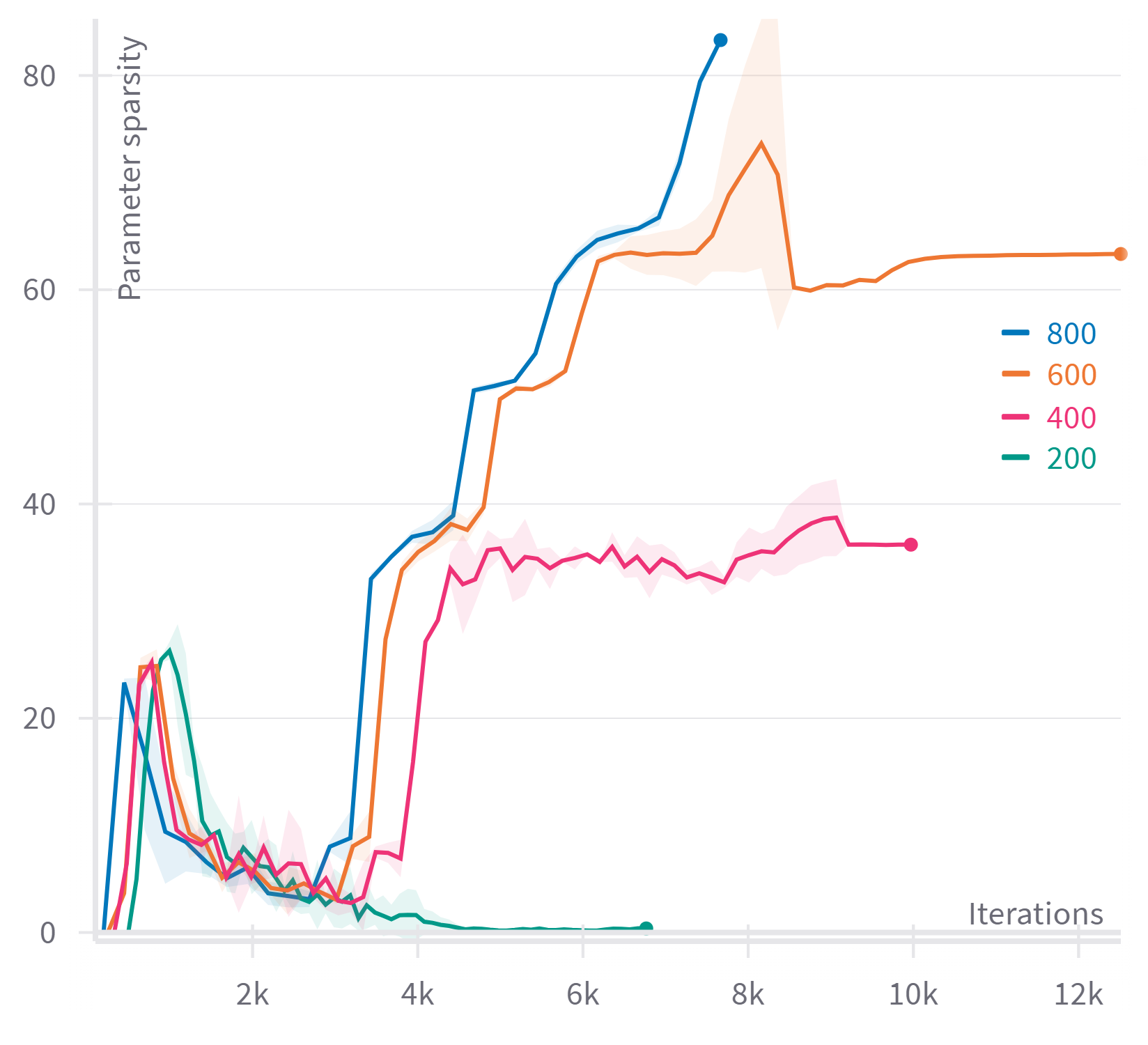}
    \caption{The graph represents the progression of parameter sparsity during training for different training data sizes for edge detection of the NYU-v2 dataset. For the rest of the task, a similar pattern can be observed.}
    \label{fig:sp_vs_data}
  \end{minipage}\hfill
  \begin{minipage}{0.53\textwidth}
    \begin{subfigure}[b]{0.45\linewidth}
      \includegraphics[width=\linewidth]{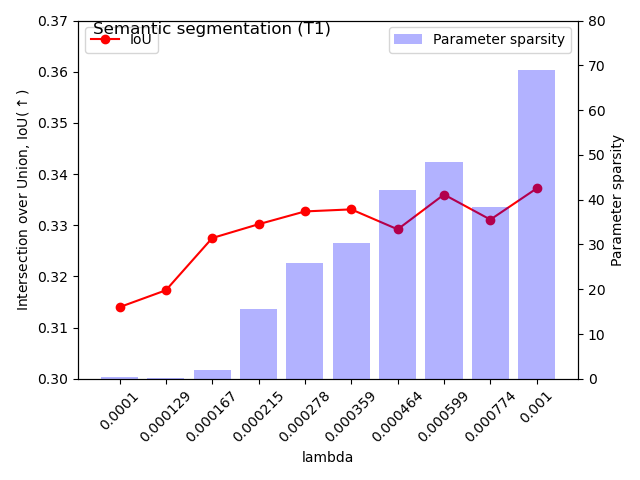}
    \end{subfigure}
    \begin{subfigure}[b]{0.45\linewidth}
      \includegraphics[width=\linewidth]{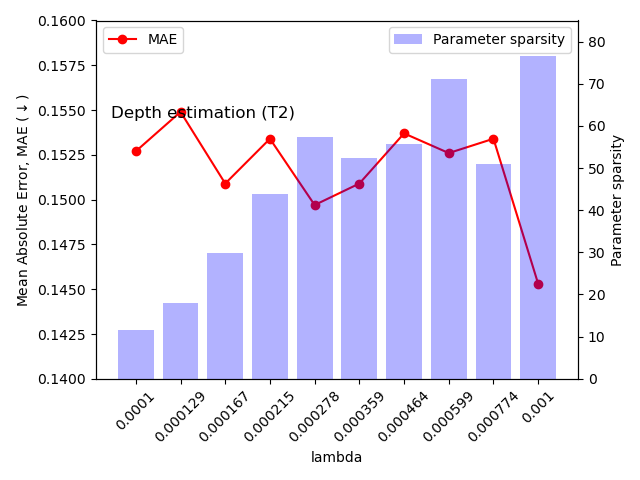}
    \end{subfigure}\\
    \begin{subfigure}[b]{0.45\linewidth}
      \includegraphics[width=\linewidth]{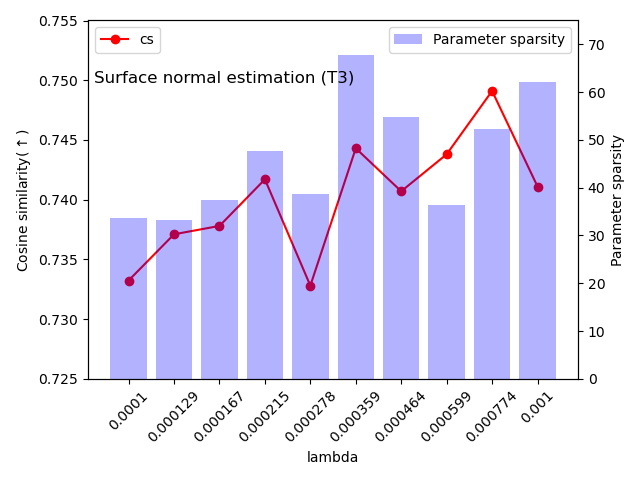}
    \end{subfigure}
    \begin{subfigure}[b]{0.45\linewidth}
      \includegraphics[width=\linewidth]{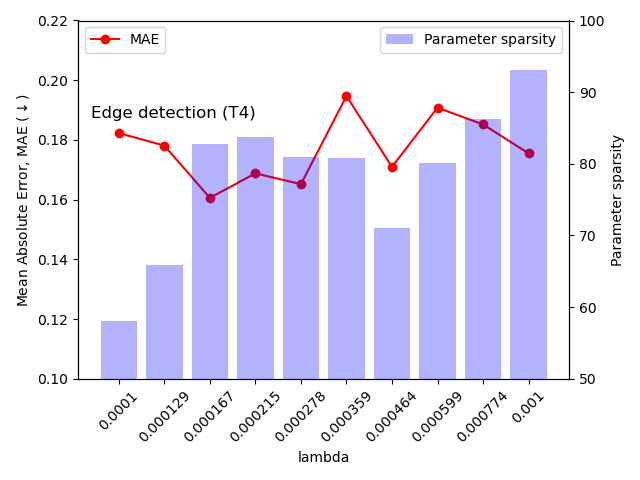}
    \end{subfigure}
    \captionof{figure}{The plots illustrate the impact of the regularization parameter $\lambda$ on parameter sparsity (in \%) and the single-task performance metrics across four tasks within the NYU-v2 dataset.}
    \label{fig:figures}
  \end{minipage}
 \end{figure} 
\subsection{Ablation studies}
\textbf{Dynamics of sparsity relative to dataset size}:
Fig.~\ref{fig:sp_vs_data} illustrates the relation between dataset size and parameter sparsity during the model training. 
A larger dataset size appears to lead to higher sparsity, which means that more data allows the model to effectively identify and sparsify irrelevant features, potentially leading to better representation learning.
Therefore, it is necessary to carry out phases 1 and 2 of this study using datasets of the same size to maintain consistency. 
This is due to the observation that varying dataset sizes lead to different sparsity patterns, which in turn influence the selection of the most suitable layer for a task for a specific dataset size.
Additionally, suppose we observe the convergence behavior of the parameter sparsity during training. In that case, initially, there is some variability, but as the training progresses, the model becomes more confident about which parameters to keep and which to prune. 
With a smaller dataset size(200) where the model displays no sparsity, the \ac{LOMT} model will resemble dense \ac{MTL} configurations, where task heads receive input from the full spectrum of the backbone network's features.

\noindent
\textbf{Choosing $\lambda$ for sparse \ac{STL} \ie phase 1}: 
The sparsity parameter $\lambda$ regulates the intensity of sparsity within a model. 
Initial experiments showed that any value of $\lambda < 0.0001$ did not induce sparsity at all, while for $\lambda > 0.001$, the model reached a very high sparsity around 90-95\% in just 2-3 epochs, not allowing the parameters to be trained sufficiently.
To find an appropriate parameter value, we conducted an exponential grid search between the values of $0.0001 \leq \lambda \leq 0.001$.
We employed an exponential grid search because the impact of $\lambda$ on sparsity is not necessarily linear; small changes may have significant effects.
After analyzing the effects of varying $\lambda$ values on model performance and amount of parameter sparsity (see Fig.~\ref{fig:figures}), the value of 0.001 emerges as the most suitable.
It can be observed from the plots that $\lambda = 0.001$ achieves an optimal balance; it attains significant sparsity while maintaining or improving performance metrics across tasks. 
By optimal sparsity, we mean that the sparsity at $\lambda = 0.001$ is high enough to benefit from the advantages of a sparse model but not so significantly high that it severely degrades the performance.
Therefore, for the sparse \ac{STL} experiments, we fixed the values of the sparsity parameter $\lambda$ to 0.001.

\section{Conclusion and future scope} \label{sec:conclusion}

The effectiveness of the proposed \ac{LOMT} models is verified through a comprehensive performance analysis, which includes both quantitative and qualitative evaluations presented in this work. 
In conclusion, \ac{LOMT} models excel across various task combinations, demonstrating their effectiveness in leveraging group sparsity to identify and utilize task-relevant layers.
Therefore, by carefully selecting which layers to activate for each task, the tailored \ac{LOMT} models create a focused learning process that enhances task-specific performance while also benefiting from shared representation learning.
A notable aspect of the proposed approach is modularity since it allows for easy incorporation of new tasks into the pre-existing multi-task network; this adaptability, however, comes with the challenge of layer selection using sparse single-task networks prior to \ac{MTL}.

Nevertheless, our work is not without limitations. 
The selection of optimal layers is highly dependent on the initial single-task sparsity induction, which may not always accurately capture the subtle interconnections between tasks in a multi-task setting. 
In addition, although \ac{LOMT} models have shown improved performance in many cases, they may encounter difficulties when tasks share contradictory features, and the negative transfer effect becomes more prominent. 
Future work should focus on improving the process of identifying relevant layer,  potentially by employing more dynamic and adaptive techniques that consider the interaction across tasks. 
Furthermore, by applying our technique to a wider variety of tasks and network architectures, we can further confirm the versatility and robustness of \ac{LOMT} models.

\section*{Acknowledgement}
The authors would like to express their gratitude for the supercomputing resource Berzelius provided by National Supercomputer Centre at Linköping University and the Knut and Alice Wallenberg foundation.
Additionally, we would like to thank all the reviewers who have contributed their invaluable feedback to improve this manuscript. 

\bibliographystyle{splncs04}
\bibliography{mybibliography}
\end{document}